%% file: main.tex
\documentclass[10pt,twocolumn,letterpaper]{article}

\usepackage{cvpr}
\usepackage{times}
\usepackage{epsfig}
\usepackage{graphicx}
\usepackage{subcaption}
\usepackage{amsmath}
\usepackage{amssymb}

\newcommand{\needtocheck}[1]{{\color{red}{#1}}} 

\usepackage[font={small}]{caption}
\usepackage{multirow}
\usepackage{ifthen}
\usepackage{bm}
\usepackage[pagebackref=true,breaklinks=true,letterpaper=true,colorlinks,bookmarks=false]{hyperref}
\cvprfinalcopy 

\def\cvprPaperID{740} 

\begin{document}

\title{\vspace{-0.8cm}Deep Network Interpolation \\ for Continuous Imagery Effect Transition}
\vspace{-0.3cm}
\author{
	Xintao Wang$^{1}$ \hspace{9pt} Ke Yu$^{1}$ \hspace{9pt} Chao Dong$^{2}$ \hspace{9pt} Xiaoou Tang$^{1}$ 
	\hspace{9pt}Chen Change Loy$^{3}$\\
	\vspace{-0.15cm}
	\small{$^{1}$CUHK - SenseTime Joint Lab, The Chinese University of Hong Kong} \\
	\vspace{-0.15cm}
	\small{$^{2}$Shenzhen Institutes of Advanced Technology, Chinese Academy of Sciences} \\
	\vspace{-0.15cm}
	\small{$^{3}$Nanyang Technological University, Singapore}\\
	{\tt\small \{wx016, yk017, xtang\}@ie.cuhk.edu.hk \hspace{5pt} chao.dong@siat.ac.cn\hspace{5pt} 
	ccloy@ntu.edu.sg}\\
}

\newboolean{putfigfirst}

\setboolean{putfigfirst}{true}
\ifthenelse{\boolean{putfigfirst}}{

\twocolumn[{%
	\renewcommand\twocolumn[1][]{#1}%
	\vspace{-1em}
	\maketitle
	\vspace{-1em}
	\begin{center}
		\centering 
		\vspace{-0.3in}
		\includegraphics[width=\linewidth]{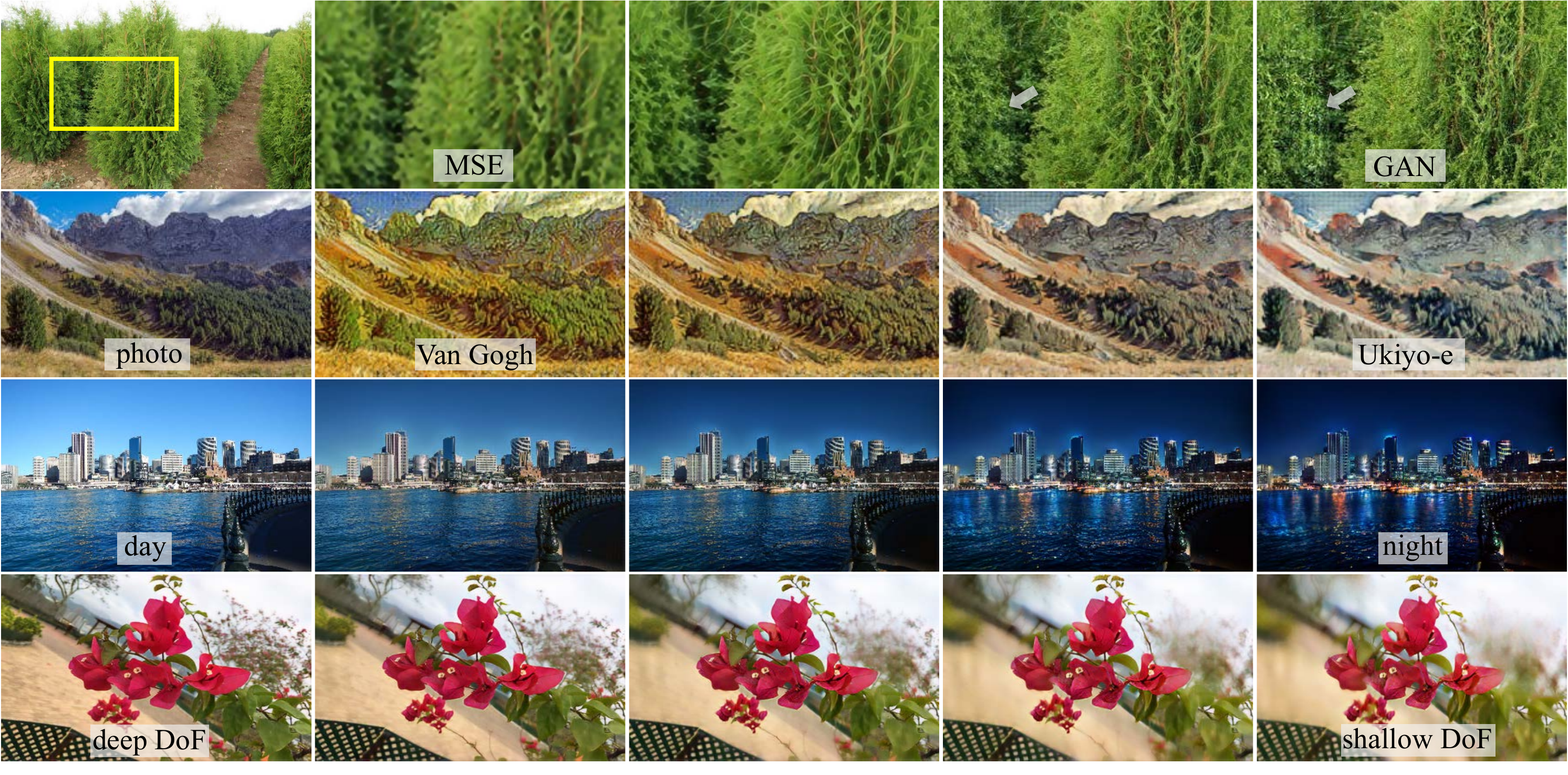}
		\vspace{-0.7cm}
		\captionof{figure}{Deep network interpolation is capable of generating continuous imagery effect transitions. 
		(\textit{$1st$ row}) 
		from MSE effect to GAN effect in super-resolution; (\textit{$2nd$ row}) from Van Gogh style to Ukiyo-e 
		style; (\textit{$3rd$ row}) from day photo to night one; (\textit{$4th$ row}) from deep depth of field 
		(DoF) to shallow one. More applications are provided in Sec.~\ref{sec:applications}. (\textbf{Zoom in for best 
		view})}
		\label{fig:teaser}
	\end{center}%
}]
}
{
\maketitle
\thispagestyle{empty}
}

\begin{abstract}
\vspace{-0.3cm}
Deep convolutional neural network has demonstrated its capability of learning a deterministic mapping for the 
desired imagery effect. However, the large variety of user flavors motivates the possibility of continuous transition 
among different output effects.
Unlike existing methods that require a specific design to achieve one particular transition (\eg, style transfer), we 
propose a simple yet universal approach to attain a smooth control of diverse imagery effects in many low-level vision 
tasks, including image restoration, image-to-image translation, and style transfer. 
Specifically, our method, namely Deep Network Interpolation (DNI), applies linear interpolation in the parameter space 
of two or more correlated networks. A smooth control of imagery effects can be achieved by tweaking the 
interpolation coefficients. In addition to DNI and its broad applications, we also investigate the mechanism of network 
interpolation from the perspective of learned filters.

\end{abstract}

\input{sections/1_introduction}
\input{sections/2_related_work}
\input{sections/3_method}
\input{sections/4_experiments}

\section{Conclusion}
In this paper, we propose a novel notion of interpolation in the parameter space, \ie, applying linear interpolation 
among the corresponding parameters of multiple correlated networks. The imagery effects change smoothly while adjusting 
the interpolation coefficients. With extensive experiments on super-resolution, denoising, image-to-image translation 
and style transfer, we demonstrate that the proposed method is applicable for a wide range of low-level vision tasks 
despite its simplicity. Compared with existing methods that achieve continuous transition by task-specific designs, 
our method is easy to generalize with negligible computational overhead.
Future work will investigate the effects of network interpolation on high-level tasks.

\clearpage
\clearpage
{\small
\bibliographystyle{ieee}
\bibliography{short,bib}
}

\clearpage
\appendix
\input{sections/5_appendix}

\end{document}

%% file: sections/1_introduction.tex

\section{Introduction}
\ifthenelse{\boolean{putfigfirst}}
{}
{
\begin{figure*}[t]
	\begin{center}
		\includegraphics[width=\linewidth]{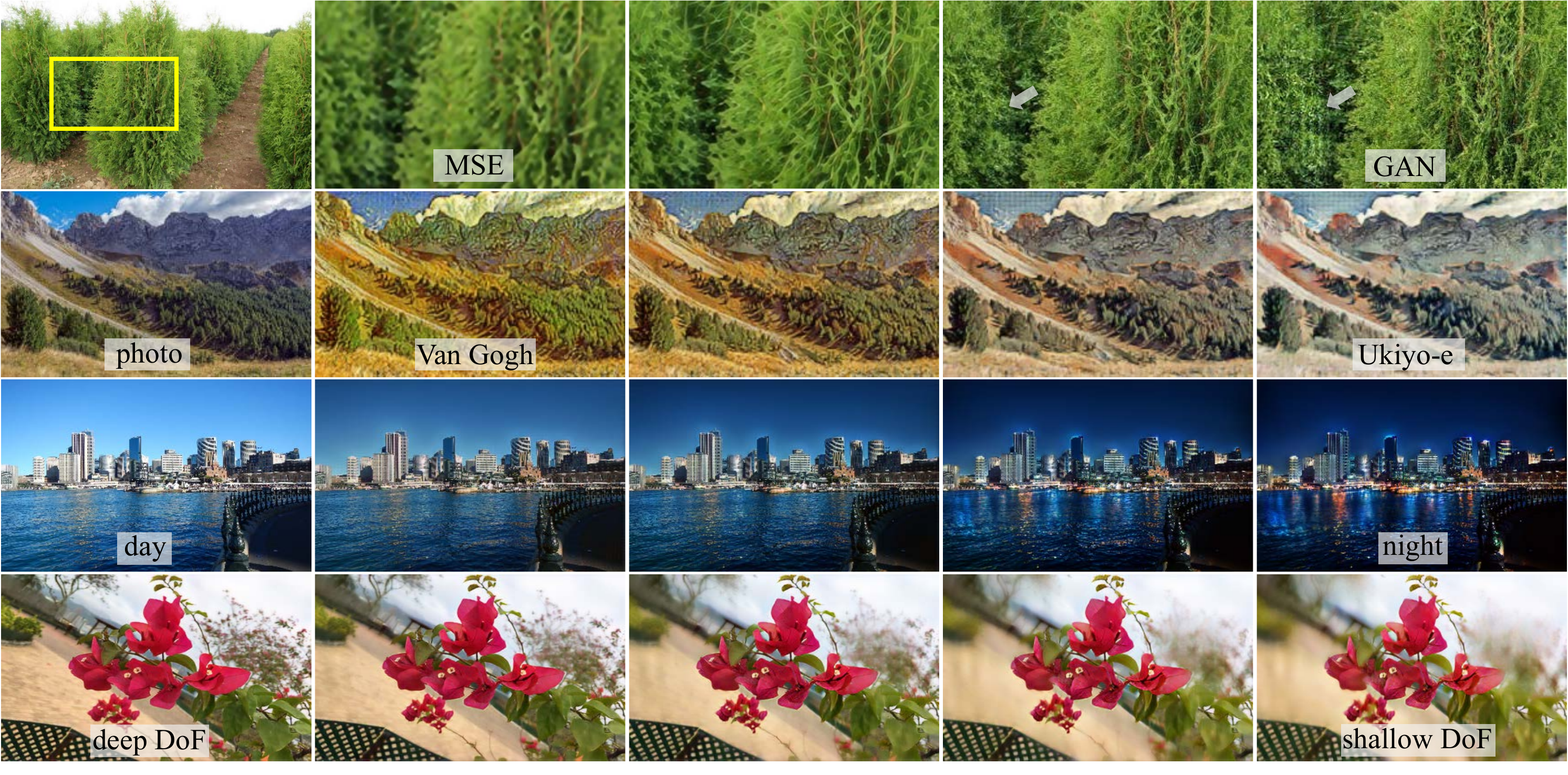}
		\caption{DNI is capable of generating continuous imagery effect transitions. (\textit{$1st$ row}) 
			from MSE effect to GAN effect in super-resolution; (\textit{$2nd$ row}) from Van Gogh style to C\'ezanne 
			style; (\textit{$3rd$ row}) from day photo to night one; (\textit{$4th$ row}) from deep depth of field 
			(DoF) to shallow one. More applications are provided in Sec.~\ref{sec:applications}. (\textbf{Zoom in for 
			best view})}
		\label{fig:teaser}
	\end{center}
	\vspace{-0.5cm}
\end{figure*}
}

Deep convolutional neural (CNN) network has achieved a great success in many low-level vision tasks, such as image 
restoration~\cite{dong2014learning,kim2016accurate,ledig2017photo,burger2012image,zhang2017beyond,dong2015compression,nah2017deep},
image-to-image translation~\cite{isola2017image,wang2017high,liu2017unsupervised,zhu2017unpaired} and image style 
transfer~\cite{gatys2016image,johnson2016perceptual,dumoulin2016learned}.
For each specific task, the deep network learns a deterministic mapping and outputs a fixed image for the same inputs. However, one determined output is unable to satisfy diverse user flavors and meet the needs in various 
scenarios, limiting the applicability for practical use.

In many real-world applications, it is desired to have a smooth control for continuous transition among different 
output effects.
For instance, 
1) in super-resolution, models trained with the mean-square-error (MSE) loss~\cite{timofte2017ntire} tend to produce 
over-smooth images while those with the generative adversarial network (GAN)~\cite{ledig2017photo} generate vivid 
details but with some unpleasant noise (\eg, Fig~\ref{fig:teaser}, $1st$ row).
A balanced result between these two different effects would be more visual-pleasing with reduced artifacts.
2) Many image restoration tasks deal with multiple degradation levels, such as different noise levels and blur kernels.
Most existing methods can only handle limited degradation levels. It is costly to train lots of models for continuous 
degradation levels in practice. Thus, a model with the flexibility of adjusting the restoration strength would 
expand the application coverage.
3) In artistic manipulation like image-to-image translation and image style transfer, different users have 
different aesthetic flavors. Achieving a smooth control for diverse effects with a sliding bar are appealing in 
these applications.

Several approaches have been proposed to improve the CNN's flexibility for producing continuous transitions in 
different tasks.
Take image style transfer as an example, adaptive scaling and shifting parameters are used in instance 
normalization layers~\cite{dumoulin2016learned,huang2017arbitrary} for modeling different styles.
Interpolating these normalization parameters for different styles produces the combination of various artistic styles.
In order to further control the stroke size in the stylized results, a carefully-designed pyramid structure consisting 
of several stroke branches are proposed~\cite{jing2018stroke}.
Though these methods are able to realize continuous transition, there are several drawbacks:
1) These careful designs are problem-specific solutions, lacking the generalizability to other tasks.
2) Modifications to existing networks are needed, thus complicate the training process.
3) There is still no effective and general way to solve the smooth control in tasks like balancing MSE and GAN effects in 
super-resolution.

In this paper, we address these drawbacks by introducing a more general, simple but effective approach, known as \textit{Deep 
Network Interpolation} (DNI). Continuous imagery effect transition is achieved via linear interpolation in the \textit{parameter space} of existing trained networks.
Specifically, provided with a model for a particular effect $\mathcal{A}$, we fine-tune it to realize another relevant 
effect $\mathcal{B}$. 
DNI applies linear interpolation for all the corresponding parameters of these two deep networks. Various interpolated 
models can then be derived by a controllable interpolation coefficient. Performing feed-forward operations on these 
interpolated models using the same input allows us to outputs with a continuous transition between the different 
effects $\mathcal{A}$ and $\mathcal{B}$.
%

Despite its simplicity, the proposed DNI can be applied to many low-level vision tasks. Some examples are presented 
in Fig~\ref{fig:teaser}. Extensive applications showcased in Sec.~\ref{sec:applications} demonstrate that deep network interpolation 
is \textit{generic} for many problems.
DNI also enjoys the following merits.
1) The transition effect is \textit{smooth} without abrupt changes during interpolation. The transition can be easily controlled by an interpolation coefficient.
2) The linear interpolation operation is \textit{simple}. No network training is needed for each transition and the computation for DNI is 
negligible.
3) DNI is \textit{compatible} with popular network structures, such as VGG~\cite{simonyan2014very}, 
ResNet~\cite{he2016deep} and DenseNet~\cite{huang2016densely}.
%

Our main contribution in this work is the novel notion of interpolation in parameter space, and its application in low-level vision tasks. 
%
%
We demonstrate that interpolation in the parameter space could achieve much better results than mere pixel interpolation. 
We further contribute a systematic study that investigates the mechanism and effectiveness of parameter interpolation through carefully analyzing the filters learned. 

%% file: sections/2_related_work.tex

\section{Related Work}
\noindent\textbf{Image Restoration.}
CNN-based approaches have led to a series of breakthroughs for several image restoration tasks including 
super-resolution~\cite{dong2014learning,kim2016accurate,lim2017enhanced,lai2017deep,tai2017image,ledig2017photo}, 
denoising~\cite{burger2012image,zhang2017beyond}, de-blocking~\cite{wang2016d3,galteri2017deep} and 
deblurring~\cite{xu2014deep,sun2015learning,nah2017deep}.
While most of the previous works focus on addressing one type of distortion without the flexibility of adjusting the 
restoration strength, there are several pioneering works aiming to deal with various practical scenarios with 
controllable ``hyper-parameters''.
Zhang \etal~\cite{zhang2017learning} adopt CNN denoisers to solve image restoration tasks by manually selecting the 
hyper-parameters in a model-based optimization framework.
However, a bank of discriminative CNN denoisers are required and the hyper-parameter selection in optimization is not a 
trivial task~\cite{dong2013nonlocally}.
SRMD~\cite{zhang2018learning} proposes an effective super-resolution network handling multiple degradations by taking 
an degradation map as extra inputs. However, the employed dimensionality stretching strategy is problem-specific, 
lacking the generalizability to other tasks.

\noindent\textbf{Image Style Transfer.}
Gatys \etal~\cite{gatys2016image} propose the neural style transfer algorithm for artistic stylization. A number of 
methods are developed to further improve its performance and 
speed~\cite{ulyanov2016texture,johnson2016perceptual,li2016combining}.
In order to model various/arbitrary styles in one model, several techniques are developed, including conditional 
instance normalization~\cite{dumoulin2016learned}, adaptive instance normalization 
\cite{huang2017arbitrary,ghiasi2017exploring} and whitening and coloring 
transforms~\cite{li2017universal}.
These carefully-designed approaches are also able to achieve user control.
For example, interpolating the normalization parameters of different styles produces the combination of various 
artistic styles. The balance of content and style can be realized by adjusting the weights of their corresponding 
features during mixing. In order to control the stroke size in the stylized results, a specially designed pyramid structure consisting of 
several stroke branches are further proposed~\cite{jing2018stroke}. 
In these studies, different controllable factors require specific structures and strategies.

\noindent\textbf{Image-to-image Translation.}
Image-to-image translation~\cite{isola2017image,wang2017high,liu2017unsupervised,zhu2017unpaired,huang2018multimodal} 
aims at learning to translate an image from one domain to another. 
For instance, from landscape photos to Monet paintings, and from smartphone snaps to professional DSLR photographs. 
These methods can only transfer an input image to a specific target manifold and they are unable to produce continuous 
translations. 
The controllable methods proposed in image restoration and image style transfer are problem-specific and cannot be 
directly applied to the different image-to-image translation task. On the contrary, the proposed DNI is capable of dealing with all 
these problems in a general way, regardless of the specific characteristics of each task. 

\noindent\textbf{Interpolation.}
Instead of performing parameter interpolation, one can also interpolate in the pixel space or feature space.
However, it is well known that interpolating images pixel by pixel introduces ghosting artifacts since natural images 
lie on a non-linear manifold~\cite{weinberger2006unsupervised}.
Upchurch \etal~\cite{upchurch2017deep} propose a linear interpolation of pre-trained deep convolutional features 
to achieve image content changes. This method requires an optimization process when inverting the features back to the 
pixel space. Moreover, it is mainly designed for transferring facial attributes and not suitable for generating 
continuous transition effects for low-level vision tasks.
More broadly, several CNN operations in the input and feature space have been proposed to increase the model's 
flexibility.
Concatenating extra conditions to inputs~\cite{zhang2018learning} or to middle features~\cite{li2017diversified} alters 
the network behavior in various scenes.
Modulating features with an affine transformation~\cite{perez2017film,dumoulin2018feature-wise,wang2018sftgan} is able 
to effectively incorporate other information.
Different from these works, we make an attempt to investigate the manipulation in the parameter space.


%% file: sections/3_method.tex

\section{Methodology}

\subsection{Deep Network Interpolation}
Many low-level vision tasks, \eg, image restoration, image style transfer, and image-to-image translation, 
aim at mapping a corrupted image or conditioned image $x$ to the desired one $y$.
Deep convolutional neural networks are applied to directly learn this mapping function $G_{\theta}$ 
parametrized by $\theta$ as $y=G_{\theta}(x)$.

Consider two networks $G^{A}$ and $G^{B}$ with the same structure, achieving different effects 
$\mathcal{A}$ and $\mathcal{B}$, respectively.
The networks consist of common operations such as convolution, up/down-sampling and non-linear activation.
The parameters in CNNs are mainly the weights of convolutional layers, called \textit{filters}, filtering 
the input image or the precedent features.
We assume that their parameters $\theta_{A}$ and $\theta_{B}$ have a ``strong correlation'' with each 
other, \ie, the filter orders and filter patterns in the same position of $G^{A}$ and $G^{B}$ are similar. 
This could be realized by some constraints like fine-tuning, as will be analyzed in 
Sec.~\ref{subsec:filter_analysis}. This assumption provides the possibility for meaningful interpolation.

Our aim is to achieve a continuous transition between the effects $\mathcal{A}$ and $\mathcal{B}$. We do so 
by the proposed Deep Network Interpolation (DNI). 
DNI interpolates all the corresponding parameters of these two models to derive a new interpolated model 
$G^{interp}$, whose parameters are:
\begin{equation}
	\theta_{interp} = \alpha \ \theta_{A} + (1-\alpha) \ \theta_{B},
\end{equation}
where $\alpha \in [0, 1]$ is the interpolation coefficient. Indeed, it is a linear interpolation of the two 
parameter vectors $\theta_{A}$ and $\theta_{B}$.
The interpolation coefficient $\alpha$ controls a balance of the effect $\mathcal{A}$ and $\mathcal{B}$. By 
smoothly sliding $\alpha$, we achieve continuous transition effects without abrupt changes.

Generally, DNI can be extended for $N$ models, denoted by $G^{1}, G^{2}, ..., G^{N}$, whose parameters have 
a ``close correlation'' with each other. 
The DNI is then formulated as:
\begin{equation}\label{eq:DNI_Nmodels}
	\theta_{interp} = \alpha_{1} \theta_{1} + \alpha_{2} \theta_{2} + ... + \alpha_{N} \theta_{N},
\end{equation}
where $\alpha_{i}$ satisfy $\alpha _{i}\geq 0$ and $ \alpha _{1}+\alpha _{2}+\cdots +\alpha _{N}=1$.
In other words, it is a convex combination of the parameter vectors $\theta_{1}, \theta_{2}, 
...,\theta_{N}$.
By adjusting $(\alpha_{1}, \alpha_{2}, ..., \alpha_{N})$, a rich and diverse effects with continuous 
transitions could be realized. 

The interpolation is performed on all the layers with parameters in the networks, including convolutional 
layers and normalization layers.
Convolutional layers have two parameters, namely \textit{weights} (filters) and \textit{biases}. The biases are 
added to the results after filtering operation with filters.
Apart from the weights, DNI also operates on the biases, since the added biases influence the successive non-linear 
activation.

Batch normalization (BN) layers~\cite{ioffe2015batch} have two kinds of parameters.
1) The statistics \textit{running\_mean} and \textit{running\_var} track the mean and variance of the whole 
dataset during training and are then used for normalization during evaluation. 
2) the learned parameters $\gamma$ and $\beta$ are for a further affine transformation.
During inference, all these four parameters actually could be absorbed into the precedent or successive 
convolutional layers.
Thus, DNI also performs on normalization parameters. 
Instance normalization (IN) has a similar behavior as BN, except that IN uses instance statistics computed 
from input data in both training and evaluation. We take the same action as that for BN.
In practice, the interpolation is performed on not only the weights but also the biases and further 
normalization layers. We believe a better interpolation scheme considering the property of different kinds of 
parameters is worthy of exploiting.

It is worth noticing that the choice of the network structure for DNI is flexible, as long as the structures of models to be interpolated are kept the same.
Our experiments on different architectures show that DNI is compatible with popular network structures such as 
VGG~\cite{simonyan2014very}, ResNet~\cite{he2016deep} and DenseNet~\cite{huang2016densely}.
%
We also note that the computation of DNI is negligible. The computation is only 
proportional to the number of parameters. 

\subsection{Understanding Network Interpolation} \label{subsec:filter_analysis}
We attempt to gain more understanding on network interpolation from some empirical studies. From our experiments, we observe that:
1) Fine-tuning facilitates high correlation between parameters of different networks, providing the possibility for meaningful
interpolation.
2) Fine-tuned filters for a series of related tasks present continuous changes.
3) Our analyses show that interpolated filters could fit the actual learned filters well.
Note that our analyses mainly focus on \textit{filters} since most of the parameters in CNNs are in the 
form of filters.

We present our main observations with a representative denoising task and focus on increasing noise levels with 
N20, N30, N40, N50, and N60, where N20 denotes the Gaussian noise with zero mean and variance 20.
In order to better visualize and analyze the filters, we adopt a three-layer network similar to 
SRCNN~\cite{dong2014learning}, where the first and last convolutional layers have $9\times9$ filter size. Following the 
notion of~\cite{dong2014learning}, the first and last layer layers can be viewed as a \textit{feature extraction} and 
\textit{reconstruction} layer, respectively.

\begin{figure}[t]
	\begin{center}
		\includegraphics[width=\linewidth]{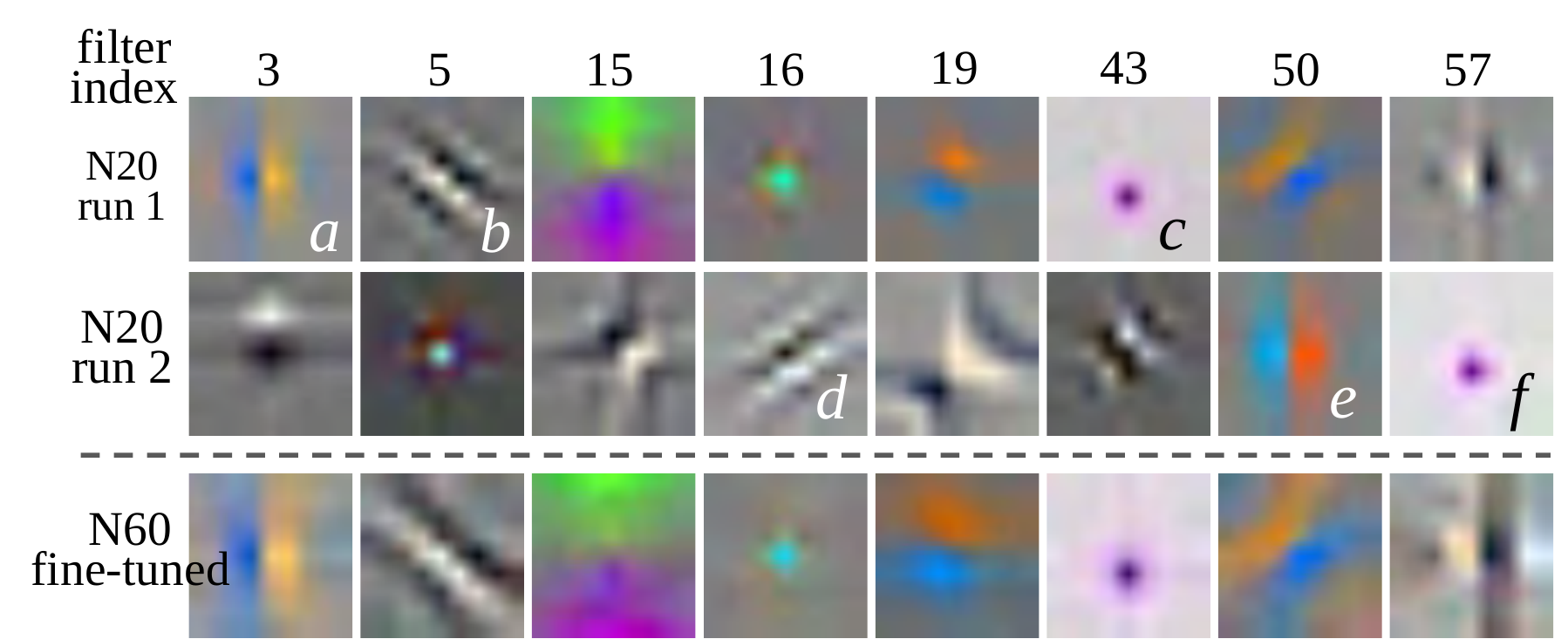}
		\vspace{-0.5cm}
		\caption{Filter correlations.
		The first two rows are the filters of different runs (both from scratch) for the denoising (N20) 
		task. The filter orders and filter patterns in the same position are different. 
		The \textit{fine-tuned} model (N60) ($3rd$ row) has a ``strong correlation'' to 
		the pre-trained one ($1st$ row).}
		\label{fig:filter_finetune}
		\vspace{-0.4cm}
	\end{center}
\end{figure}

\noindent\textbf{Fine-tuning for inter-network correlation.}
Even for the same task like denoising with the N20 level, if we simply train two models from scratch, the 
filter orders among channels and filter patterns in the corresponding positions could be very different 
(Fig.~\ref{fig:filter_finetune}). 
However, a core representation is shared between these two networks~\cite{li2015convergent}.
For instance, in Fig.~\ref{fig:filter_finetune}, filer $c$ is identical to filter $f$; filter $a$ and 
filter $e$ have a similar pattern but with different colors; filter $b$ is a inverted and rotated 
counterpart of filter $d$.

Fine-tuning, however, can help to maintain the filters's order and pattern. To show this, we fine-tune a pre-trained network (N20) to a relevant task (N60). It is observed that the filter orders and filter patterns are maintained (Fig.~\ref{fig:filter_finetune}).
The ``high correlation''  between the parameters of these two networks provides the 
possibility for meaningful interpolation. We note that besides fine-tuning, other constraints such as joint training with regularization may also achieve such inter-network correlation.

\begin{figure}[t]
	\begin{center}
		\includegraphics[width=\linewidth]{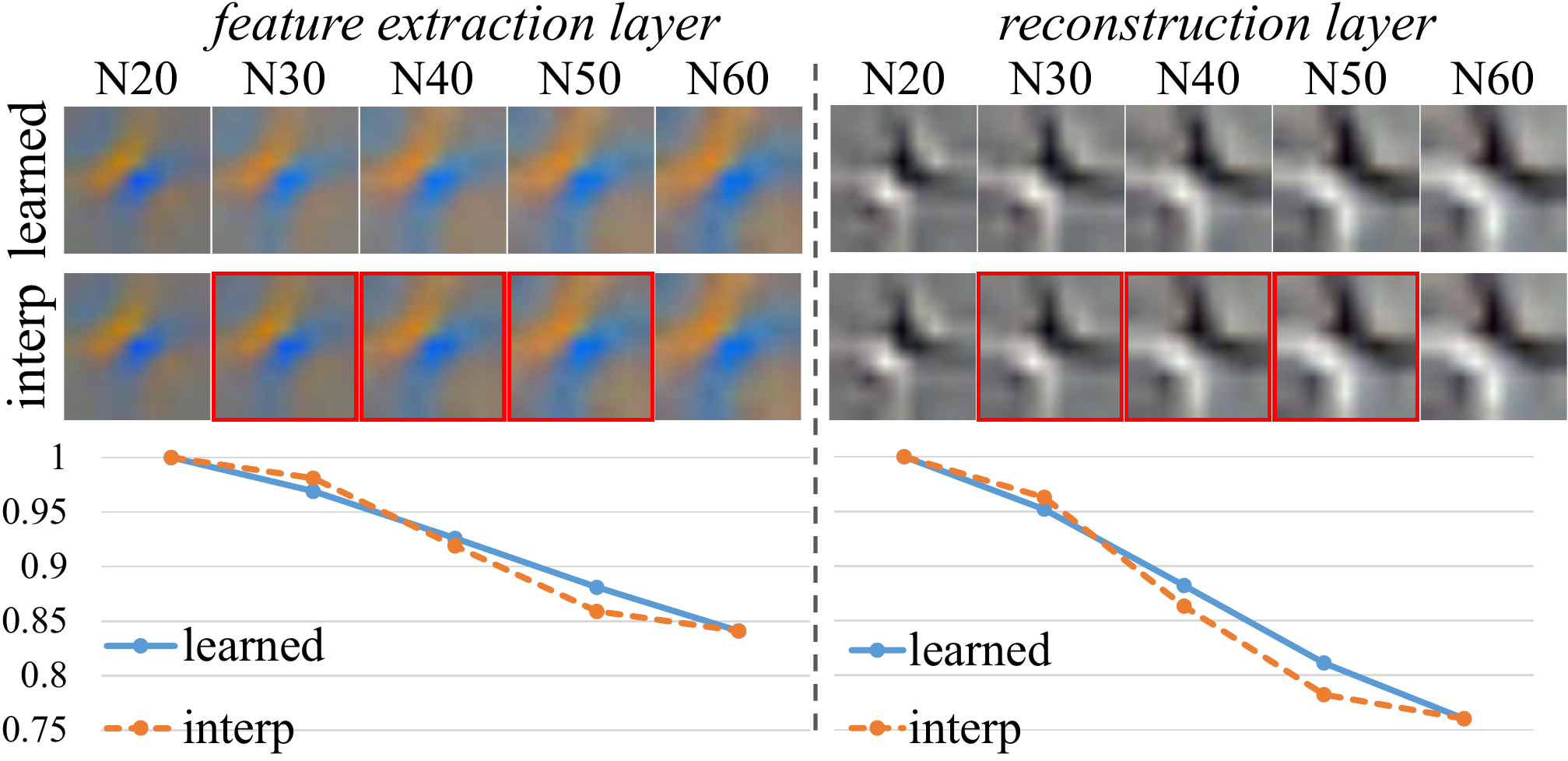}
		\vspace{-0.5cm}
		\caption{Filters with gradual changes.
			We show one representative filter for each layer.
			$\bm{1st}$ \textbf{row}: The fine-tuned filters for different noise level show gradual changes. 
			$\bm{2nd}$ \textbf{row}: The interpolated filters (with red frames) from the N20 and N60 filters could 
			visually fit those learned filters well.
			$\bm{3rd}$ \textbf{row}: The correlation curves for learned and interpolated filters are also very close.
		}
		\label{fig:filter_interp}
		\vspace{-0.4cm}
	\end{center}
\end{figure}

\noindent\textbf{Learned filters for related tasks exhibit continuous changes.}
When we fine-tune several models for relevant tasks (N30, N40, N50, and N60) from a pre-trained one (N20), the 
correspondingly learned filters have intrinsic relations with a smooth transition.
As shown in Fig.~\ref{fig:filter_interp} ($1st$ row), the trained filters show gradual changes as the noise level 
increases.
Apart from visualization, we also calculate a correlation index $\rho _{ij}$ to measure the correlations of filters 
$F_i$ and $F_j$: \vspace{-0.2cm}
\begin{equation}\label{eq:correlation_index}
	\rho _{ij} = 
	\frac{(F_i-\overline{F_i})\cdot(F_j-\overline{F_j})}
	{\sqrt{\|F_i-\overline{F_i}\|_{2}}\sqrt{\|F_j-\overline{F_j}\|_{2}}}.
\end{equation}
We choose this form (similar to the Pearson correlation) since it ignores the scale and shift influences and focus on 
the filter pattern itself.
We calculate the correlation index for each filter with the first N20 filter and plot the curve (blue curve in 
Fig.~\ref{fig:filter_interp}). The results suggest a close relationship among the learned filters, exhibiting a gradual change as 
the noise level increases. 

\noindent\textbf{The interpolated filters fit the learned filters well.}
The continuous changes of learned filters suggest that \textit{it it possible to obtain the intermediate 
filters by interpolating the two ends}.
To further verify this observation, we perform linear interpolation between the filters from the N20 and N60 models.
With optimal coefficients $\alpha$, the interpolated filters could visually fit those learned filters ($2nd$ row with 
red frames, Fig.~\ref{fig:filter_interp}).
We further calculate the correlation index for each interpolated filter with the first N20 filter.
The correlation curves for learned and interpolated filters are also very close.

The optimal $\alpha$ is obtained through the final performance of the interpolated network.
Specifically, we perform DNI with $\alpha$ from 0 to 1 with an interval of $0.05$.
The best $\alpha$ for each noise level is selected based on which that makes the interpolated network to produce the highest PSNR on the 
test dataset.

\noindent\textbf{Discussion.}
It is noteworthy that similar observations could also be found in several other tasks, such as super-resolution with 
different kernels and JPEG artifacts removal with different compression levels. We provide the details 
in the Appendix \ref{sec:filter_analyses}.

\begin{figure}[h]
	\begin{center}
		\includegraphics[width=\linewidth]{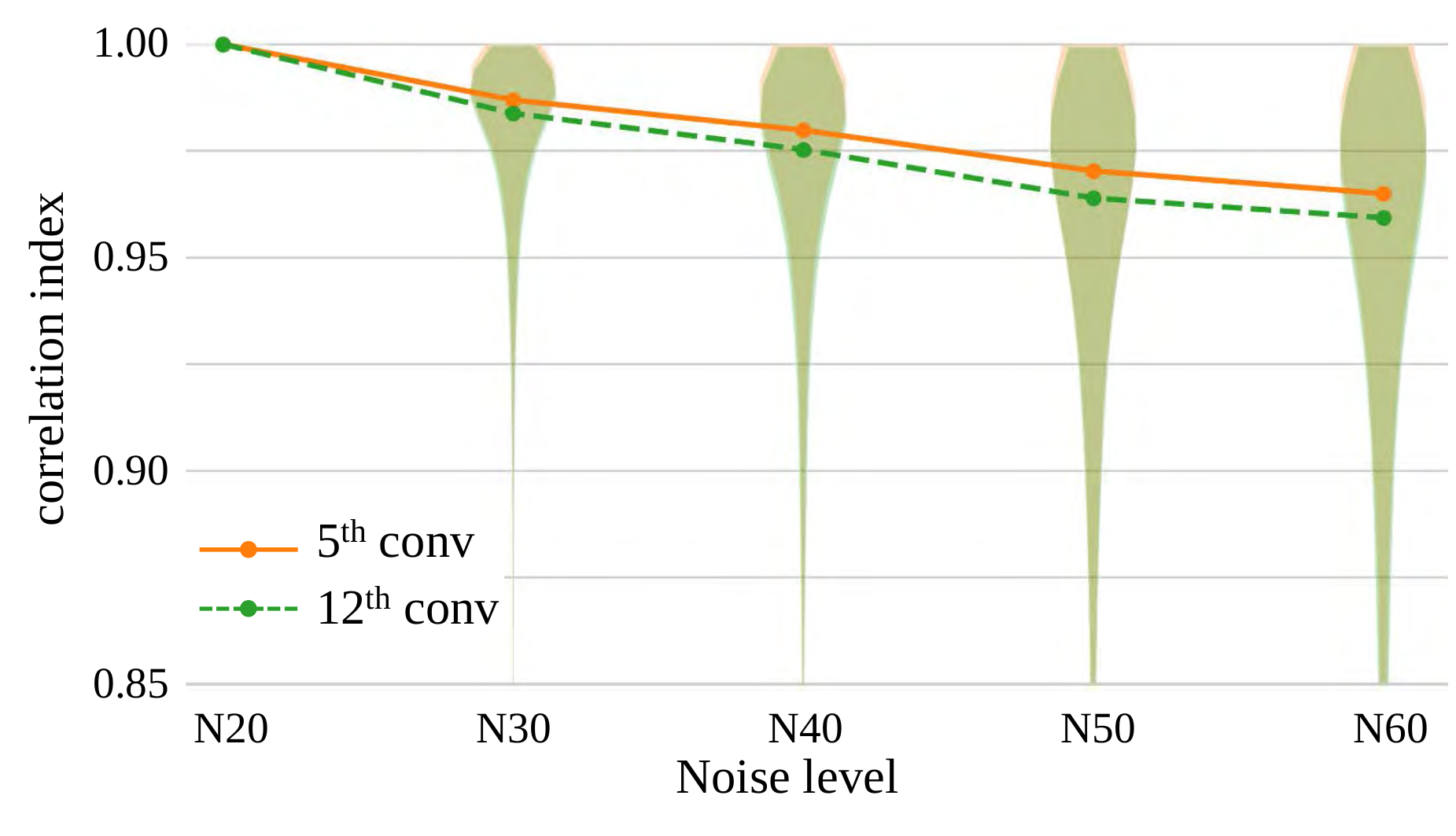}
		\caption{Filter correlation for deeper denoising networks.
			We show a front ($5th$) and a back ($12th$) layer. The curves present the median of correlation indexes and 
			the correlation distribution are also plotted. Note that the two distributions highly coincide. 
			Zoom in for subtle differences.}
		\label{fig:filter_cor_dncnn_bn}
	\end{center}
\end{figure}

The analysis above is by no means complete, but it gives a preliminary explanation behind the DNI from the filter perspective.
As the network goes deeper, the non-linearity increases and the network behaviors become more complicated.
However, we still observe a similar phenomenon for deeper networks.
Since filter visualization is difficult in deep networks, which typically designed with a stack of convolutional layers of $3\times3$ 
kernels, we adopt the correlation index (Eq.~\ref{eq:correlation_index}) to analyze the filter correlations among 
models for a series of noise levels.
We employ DnCNN~\cite{zhang2017beyond} with 17 layers and analyze the 5th (the front) and 12th (the back) convolutional 
layers. 
In Fig.~\ref{fig:filter_cor_dncnn_bn}, the correlation curves show the median of correlation indexes \wrt the first N20 
model and the correlation distribution are also plotted.
Besides the high correlation among these models, we can also observe gradual changes as the noise level increases. 
Furthermore, the front and the back convolution layers present a similar transition trend, even their distributions 
highly coincide.


%% file: sections/4_experiments.tex
\vspace{-7mm}
\section{Applications} \label{sec:applications}

We experimentally show that the proposed DNI can be applied to extensive low-level vision 
tasks, \eg, image restoration (Sec.~\ref{subsec:image_restoration}), image-to-image translation 
(Sec.~\ref{subsec:image_to_image_translation}) and image style transfer (Sec.~\ref{subsec:style_transfer}). 
Another example of smooth transitions on face attributes are presented in Sec.~\ref{subsec:semantic_transition}, 
indicating its potential for semantic changes. 
Due to space limitations, more examples and analyses are provided in the Appendix 
\ref{sec:more_applications_and_analyses} and the 
project page\footnote{\url{https://xinntao.github.io/projects/DNI}}.

\subsection{Image Restoration}
\label{subsec:image_restoration}

\begin{figure*}[th]
	\begin{center}
		\includegraphics[width=\linewidth]{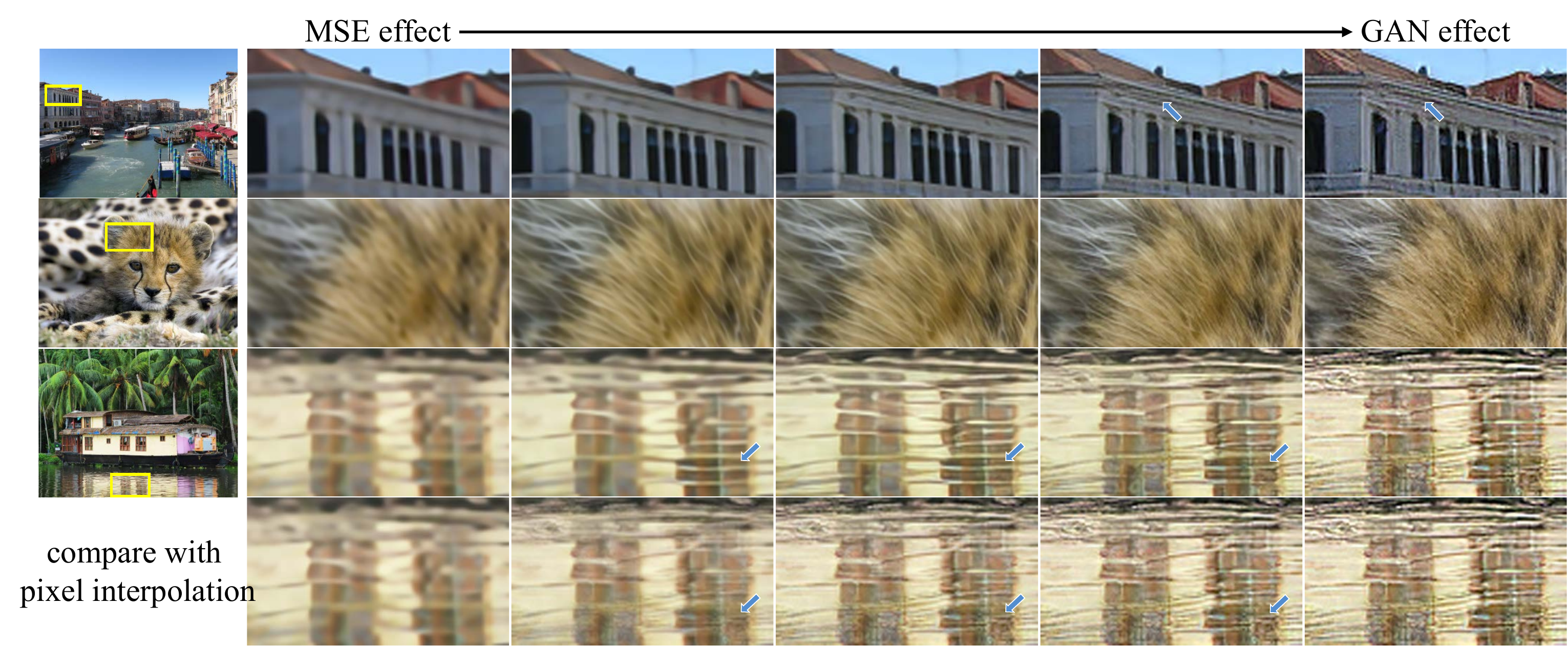}
		\caption{Balancing the MSE and GAN effects with DNI in super-resolution.
			The MSE effect is over-smooth while the GAN effect is always accompanied with unpleasant artifacts (\eg, the eaves and 
			water waves). DNI allows smooth transition from one effect to the other and produces visually-pleasing 
			results with largely reduced artifacts while maintaining the textures. In contrast, the pixel interpolation strategy fails to  separate the artifacts and textures.
			(\textbf{Zoom in for best view)}}
		\label{fig:exp_sr_interp}
	\end{center}
\end{figure*}

\noindent\textbf{Balance MSE and GAN effects in super-resolution.}
The aim of super-resolution is to estimate a high-resolution image from its low-resolution counterpart.
A super-resolution model trained with MSE loss~\cite{timofte2017ntire} tends to produce over-smooth images. We 
fine-tune it with GAN loss and perceptual loss~\cite{ledig2017photo}, obtaining results with vivid details but always 
together with unpleasant artifacts (\eg, the eaves and water waves in Fig.~\ref{fig:exp_sr_interp}).
We use dense blocks~\cite{huang2016densely,wang2018esrgan} in the network architecture and the MATLAB bicubic kernel  
with a scaling factor of 4 is adopted as the down-sampling kernel.

As presented in Fig.~\ref{fig:exp_sr_interp}, DNI is able to smoothly alter the outputs from the MSE effect to the GAN 
effect. With appropriate interpolation coefficient, it produces visually-pleasing results with largely reduced 
artifacts while maintaining the textures.
We also compare it with pixel interpolation, \ie, interpolating the output images pixel by pixel.
However, the pixel interpolation is incapable of separating the artifacts and details.
Taking the water waves for example (Fig.~\ref{fig:exp_sr_interp}), the water wave texture and artifacts
simultaneously appear and enhance during the transition. Instead, DNI first enhances the vivid water waves without 
artifacts and then finer textures and undesirable noise appears successively. The effective separation helps to 
remove the displeasing artifacts while keeping the favorable textures, superior to the pixel interpolation.
This property could also be observed in the transition of animal fur in Fig.~\ref{fig:exp_sr_interp}, where the 
main structure of fur first appears followed by finer structures and subtle textures. 

One can also obtain several models for different mixture effects, by tuning the weights of MSE loss and GAN loss during 
training. However, this approach requires tuning on the loss weights and training many networks for various balances, and 
thus it is too costly to achieve continuous control. 

\begin{figure*}[h]
	\begin{center}
		\vspace*{-4mm}
		\includegraphics[width=\linewidth]{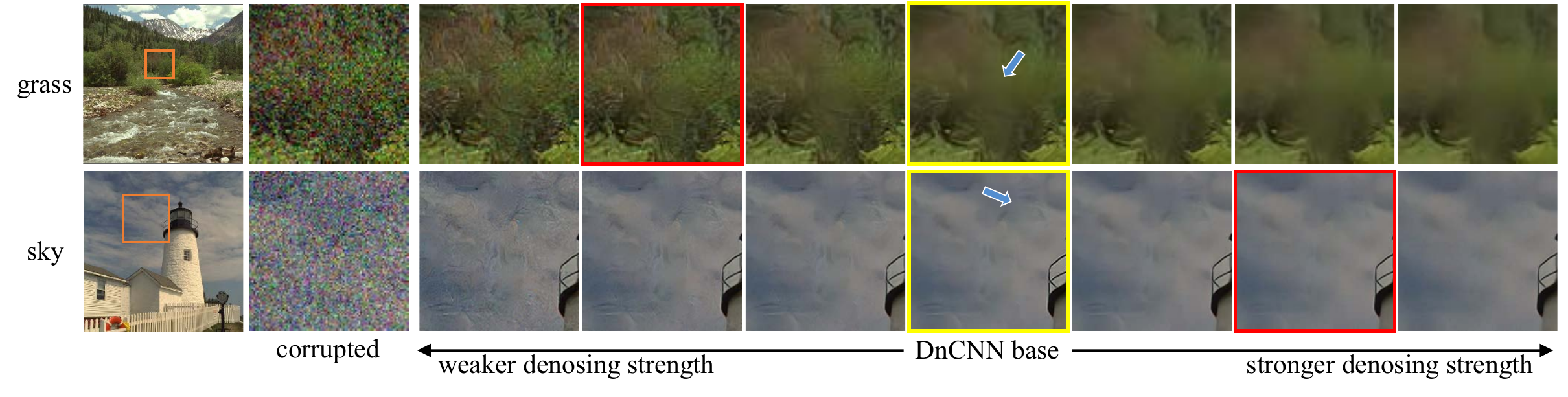}
		\caption{Adjustable denoising strength with DNI.
			One model without adjustment (with yellow frames) is unable to balance the noise removal and detail 
			preservation. For the grass, a weaker denoising strength could preserve more textures while for the sky, 
			the stronger denoising strength could obtain an artifact-free result, improving the visual quality (with 
			red frames).
			(\textbf{Zoom in for best view})}
		\label{fig:denoise_strength}
		\vspace*{-5mm}
	\end{center}
\end{figure*}
\vspace{1mm}
\noindent\textbf{Adjust denoising strength.}
The goal of denoising is to recover a clean image from a noisy observation. 
%
In order to satisfy various user demands, most popular image editing softwares (\eg, Photoshop) have controllable 
options for each tool. For example, the noise reduction tool comes with sliding bars for controlling the denoising strength and the percentage of preserving or sharpening details. 

We show an example to illustrate the importance of adjustable denoising strength. We are provided with a denoising 
model specialized in addressing a specific Gaussian noise level N40.
We use DnCNN~\cite{zhang2017beyond} as our 
\href{https://github.com/cszn/DnCNN/tree/master/TrainingCodes/dncnn_pytorch}{implementation}.
As shown in Fig.~\ref{fig:denoise_strength}, however, the determined outputs (with yellow frames) are not satisfactory due to the different imagery contents. 
In particular, the denoising strength for the grass is too strong, producing over-smooth results, while in the smooth sky region, it requires a larger strength to remove the undesired artifacts.

Existing deep-learning based approaches fail to meet this user requirements since they are trained to generate 
deterministic results without the flexibility to control the denoising strength. 
On the contrary, our proposed DNI is able to achieve adjustable denoising strength by simply tweaking the interpolation 
coefficient $\alpha$ of different denoising models for N20, N40 and N60. 
For the grass, a weaker denoising strength could preserve more details while in the sky region, the stronger 
denoising strength could obtain an artifact-free result (red frames in 
Fig.~\ref{fig:denoise_strength}).
This example demonstrates the flexibility of DNI to customize restoration results based on the task at hand and the 
specific user preference.
%

\begin{figure*}[tb]
	\centering
	\vspace*{-2mm}\begin{subfigure}[b]{0.98\linewidth}
		\includegraphics[width=\linewidth]{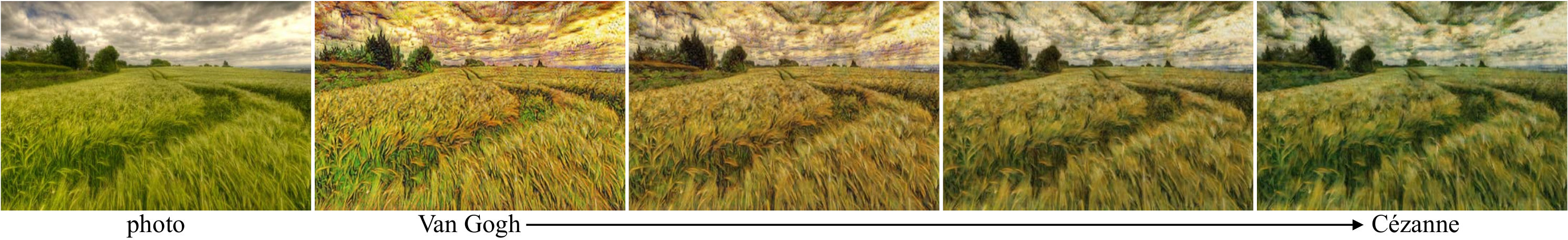}
		\vspace*{-7mm}\caption{Photos to paintings. DNI produces a smooth transition from Van Gogh's style to 
			C\'ezanne's style both in the palette and brush strokes.}
		\label{fig:vangogh_cezanne}
	\end{subfigure}
	\\
	\vspace*{0mm}\begin{subfigure}[b]{0.98\linewidth}
		\includegraphics[width=\linewidth]{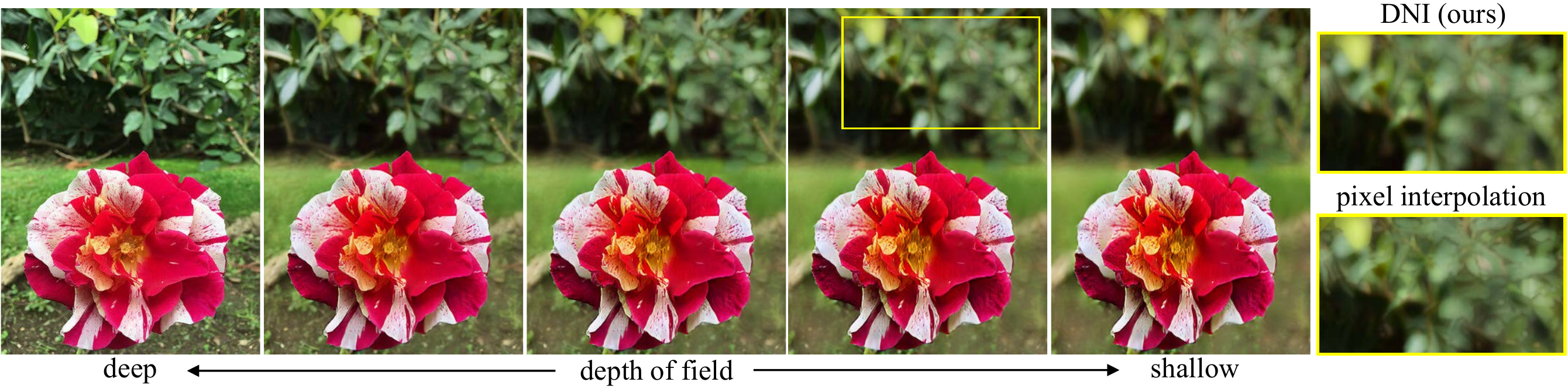}
		\vspace*{-7mm}\caption{Smooth transition on depth of field with DNI. However, pixel interpolation 
		generates ghosting artifacts. (\textbf{Zoom in for best view})}
		\label{fig:depth_of_field}
	\end{subfigure}
	\\
	\vspace*{0mm}\begin{subfigure}[b]{0.98\linewidth}
		\includegraphics[width=\linewidth]{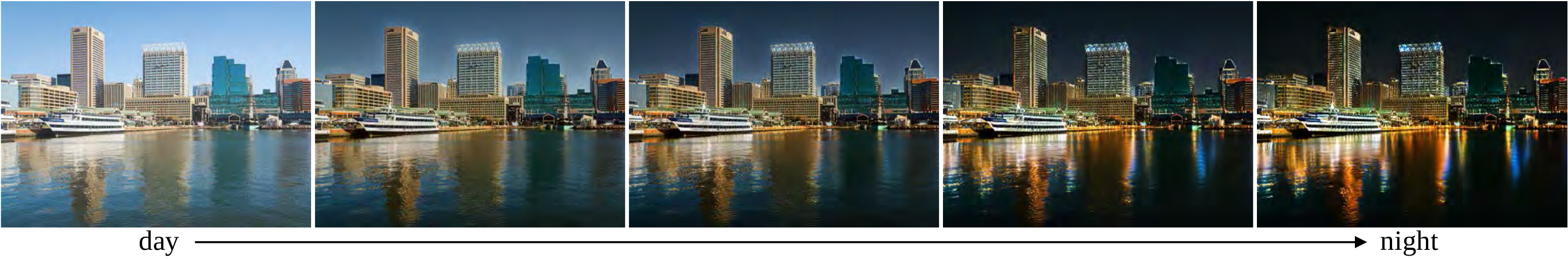}
		\vspace*{-7mm}\caption{Day photos to night ones. As the night approaches, it is 
			getting darker and the lights are gradually lit up, reflected on the water.}
		\label{fig:day_night}
	\end{subfigure}
	\vspace*{-3mm}
	\caption{Several applications for image-to-image translation. (\textbf{Zoom in for best view}) }
	\label{fig:image_to_image}
	\vspace*{-3mm}
\end{figure*}

\subsection{Image-to-image Translation}\label{subsec:image_to_image_translation}
Image-to-image translation aims at learning to translate an image from one domain to another.
Most existing approaches~\cite{isola2017image,wang2017high,liu2017unsupervised,zhu2017unpaired} can only transfer one input 
image to several discrete outputs, lacking continuous translation for diverse user flavors. 
For example, one model may be able to mimic the style of Van Gogh or C\'ezanne, but translating a landscape photo into a mixed style of these two painters is still challenging.

The desired continuous transition between two painters' styles can be easily realized by DNI.
The popular CycleGAN~\cite{zhu2017unpaired} is used as our 
\href{https://github.com/junyanz/pytorch-CycleGAN-and-pix2pix}{implementation}.
We first train a network capturing characteristics of Van Gogh, and then fine-tune it to produce 
paintings of C\'ezanne's style.
DNI is capable of generating various mixtures of these two styles given a landscape photo, by adjusting the 
interpolation coefficient. Fig.~\ref{fig:vangogh_cezanne} presents a smooth transition from Van Gogh's style to 
C\'ezanne's style both in the palette and brush strokes.
We note that DNI can be further employed to mix styles of more than two painters using Eq.~\ref{eq:DNI_Nmodels}. 
Results are provided in the Appendix 
\ref{sec:more_applications_and_analyses}. 

In addition to the translation between painting styles of a whole image, DNI can also achieve smooth and natural translation for a particular image region. 
Fig.~\ref{fig:depth_of_field} shows an example of photo enhancement to generate photos with shallower depth of field (DoF).
We train one model to generate flower photos with shallow DoF and then fine-tune it with identity mapping.
DNI is then able to produce continuous transitions of DoF by interpolating these two models.
We also compare DNI with pixel interpolation, where the results look unnatural due to the ghosting artifacts, \eg, 
translucent details appearing at the edge of blurry leaves.

DNI can be further applied to achieve continuous imagery translations in other dimensions such as light changes, \ie, transforming the day photos to night ones. 
Only trained with day and night photos, DNI is capable of generating a series of images, simulating the coming of nights. 
In Fig.~\ref{fig:day_night}, as the night approaches, it is getting darker and the lights are gradually lit up, reflected on the water.

\begin{figure*}[t]
	\begin{center}
		\includegraphics[width=0.95\linewidth]{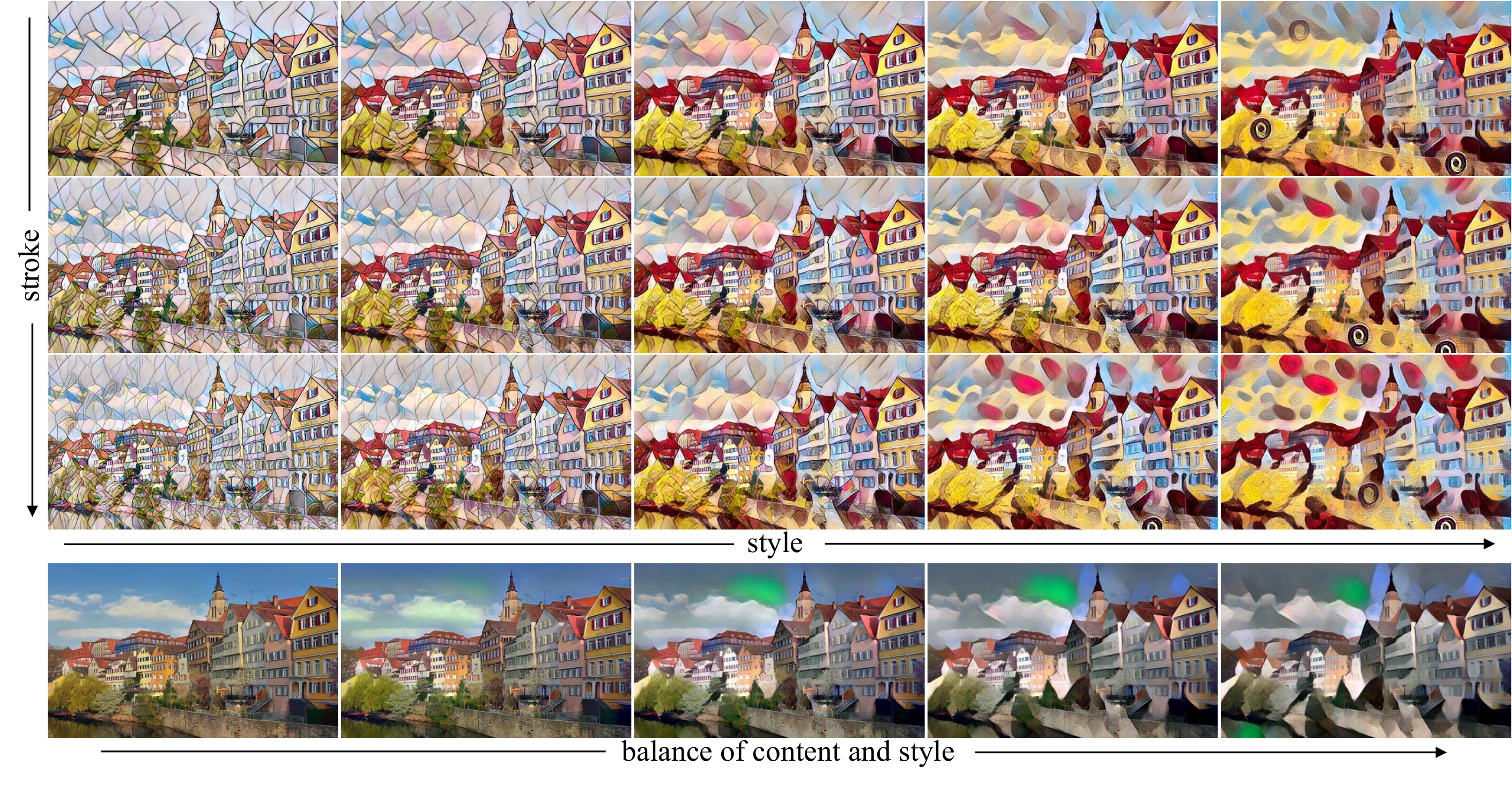}
		\caption{In image style transfer, without specific structures and strategies, DNI is capable of generating 
		smooth transitions between different styles, from large strokes to small strokes, together with balancing 
		the content and style. (\textbf{Zoom in for best view})}
		\label{fig:style_stroke_strength}
	\end{center}
\end{figure*}

\subsection{Style Transfer}\label{subsec:style_transfer}
There are several controllable factors when transferring the styles of one or many pieces of art to an input 
image, \eg, style mixture, stroke adjustment and the balance of content and style. 
Some existing approaches design specific structures to achieve a continuous control of these factors~\cite{jing2018stroke}.
On the contrary, DNI offers a general way to attain the same goals without specific solutions.
As shown in Fig.~\ref{fig:style_stroke_strength}, DNI is capable of generating smooth transitions between different 
styles, from large to small strokes, together with balancing the content and style. 
Furthermore, DNI can be applied among multiple models to achieve a continuous control of various factors simultaneously. For instance, the stroke and style can be adjusted at the same time based on user flavors, as shown in Fig.~\ref{fig:style_stroke_strength}.

\begin{figure}[t]
	\begin{center}
		\vspace{-3mm}
		\includegraphics[width=1\linewidth]{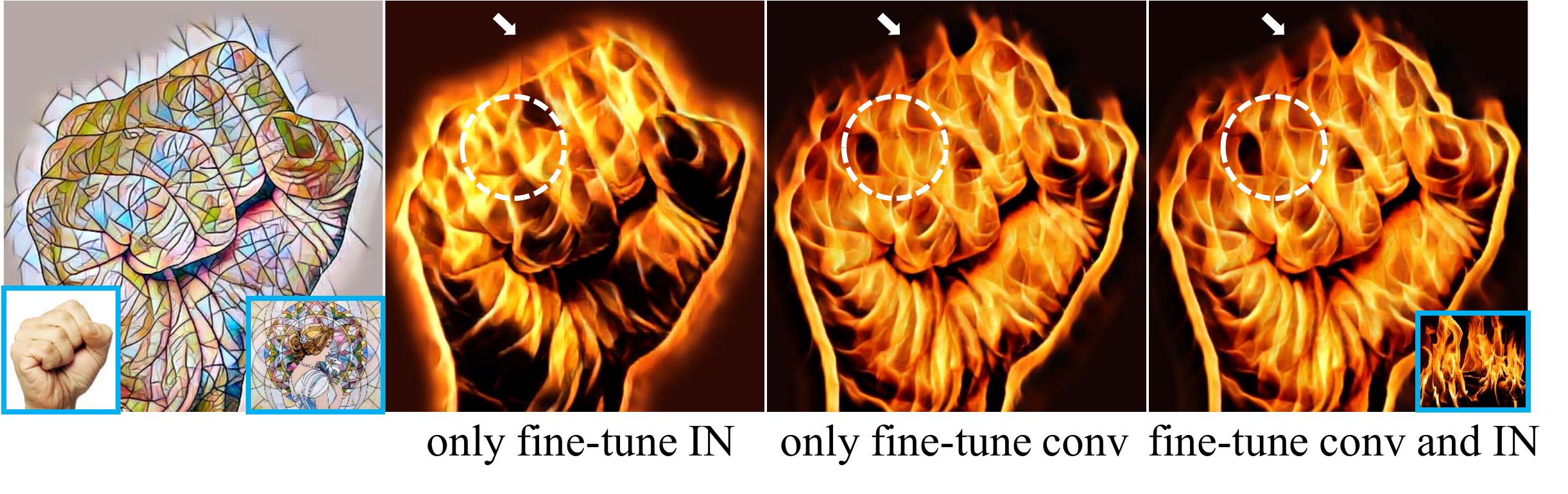}
		\caption{
		Fine-tuning only IN is effective in color transformation, however, it is unable to transfer the fire texture 
		effectively. 
		(\textbf{Zoom in for best view})}
		\label{fig:style_IN}
		\vspace*{-6mm}
	\end{center}
\end{figure}

Another branch of existing methods achieves a combination of various artistic styles by interpolating the parameters of instance normalization (IN)~\cite{dumoulin2016learned,huang2017arbitrary}. 
These approaches can be viewed as a special case of DNI, where only IN parameters are fine-tuned and interpolated. 
To clarify the difference between DNI and IN interpolation, we conduct experiments with 3 settings: 1) fine-tune IN; 2) fine-tune convolutional layers and 3) fine-tune both IN and convolutional layers.
Specifically, as shown in Fig.~\ref{fig:style_IN}, we try a challenging task to fine-tune the model from generating 
images with mosaic styles to the one with fire styles, where the two styles look very different in both 
color and texture.
It is observed that fine-tuning only IN is effective in color transformation, however, it is unable to transfer the 
fire texture effectively compared with the other two settings.
This observation suggests that convolutional layers also play an important role in style modeling, especially for the 
textures, since IN parameters may not be effective for capturing spatial information.
However, we do not claim that DNI is consistently better than IN interpolation, since IN is also effective in most 
cases. A more thorough study is left for future work.

\subsection{Semantic Transition}\label{subsec:semantic_transition}
\begin{figure}[t]
	\begin{center}
		\vspace{-3mm}
		\includegraphics[width=1\linewidth]{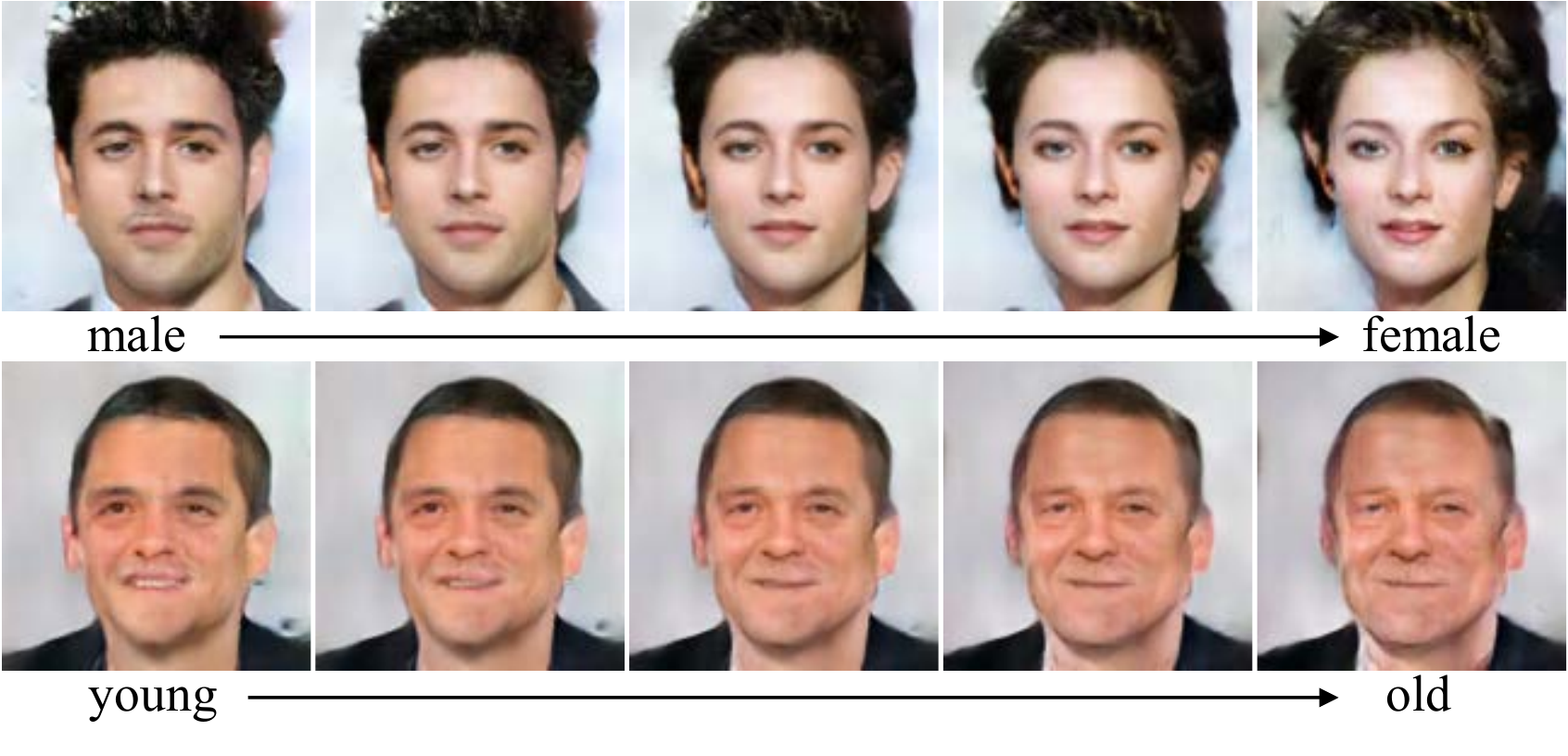}
		\caption{Smooth transitions on face attributes with DNI.}
		\label{fig:semantic_changes}
		\vspace*{-6mm}
	\end{center}
\end{figure}

Apart from low-level vision tasks, we show that DNI can also be applied for smooth transitions on face attributes, 
suggesting its potential for semantic adjustment. 
We first train a DCGAN model~\cite{radford2015unsupervised} using the CelebA~\cite{liu2015faceattributes} 
dataset with one attribute (\eg, young or male). 
After that, we fine-tune it to generate faces with another attribute (\eg, old or female).
DNI is then able to produce a series of faces with smoothly transformed attributes by interpolating these models 
(Fig.~\ref{fig:semantic_changes}).
Although neither of the interpolated models observes any data with middle attributes, the faces in middle states has 
an intermediate attribute and looks natural.

%% file: sections/5_appendix.tex

\noindent\textbf{\large{Appendix}} \label{appendix}

\vspace{0.1cm}
\textit{
	We first provide more details about filter analyses on other tasks in 
	Sec.~\ref{sec:filter_analyses}, such as super-resolution with different kernels and JPEG artifacts removal with 
	different compression qualities.
	We then provide the implementation details of deep network interpolation for different applications in  
	Sec.~\ref{sec:implementation_details}.
	Additional applications and analyses are presented in Sec.~\ref{sec:more_applications_and_analyses}. 
}

\section{Filter Analyses} \label{sec:filter_analyses}
In the main paper, we provide a preliminary explanation behind the deep network interpolation (DNI) from the filter 
perspective, through analyses of the denoising task. Specifically, the observations can be summarized as:
1) Fine-tuning facilitates high correlation between parameters of different networks, providing the possibility for 
meaningful interpolation. 2) Fine-tuned filters for a series of related tasks present continuous changes. 3) Our 
analyses show that interpolated filters could fit the actual learned filters well.

The similar observations could also be found in several other tasks, such as super-resolution with different kernels 
and JPEG artifacts removal with different compression levels. In particular, we also adopt a three-layer network 
similar to SRCNN~\cite{dong2014learning}, where the first and last convolutional layers have 9$\times$9 filter size 
(\ie, the 
same architecture as that in the main paper). Following the notion of~\cite{dong2014learning}, the first and last 
layer can be viewed as a \textit{feature extraction} and \textit{reconstruction} layer, respectively.

For super-resolution, we focus on a series of blurring kernels with 
different kernel widths, followed by a down-sampling operation. The kernel widths in our experiments are K3, K5, K7, 
K9, and K11, where K3 denotes a Gaussian blur kernel with size 3. We use the \textit{OpenCV GaussianBlur} 
function and the Gaussian kernel standard deviation can be derived from the kernel width.

For JPEG compression artifacts removal, we employ increasing compression qualities with Q10, Q20, Q30, Q40, and Q50, 
(the larger the number, the better image quality after compression).

\begin{figure}[h]
	\begin{center}
		\includegraphics[width=\linewidth]{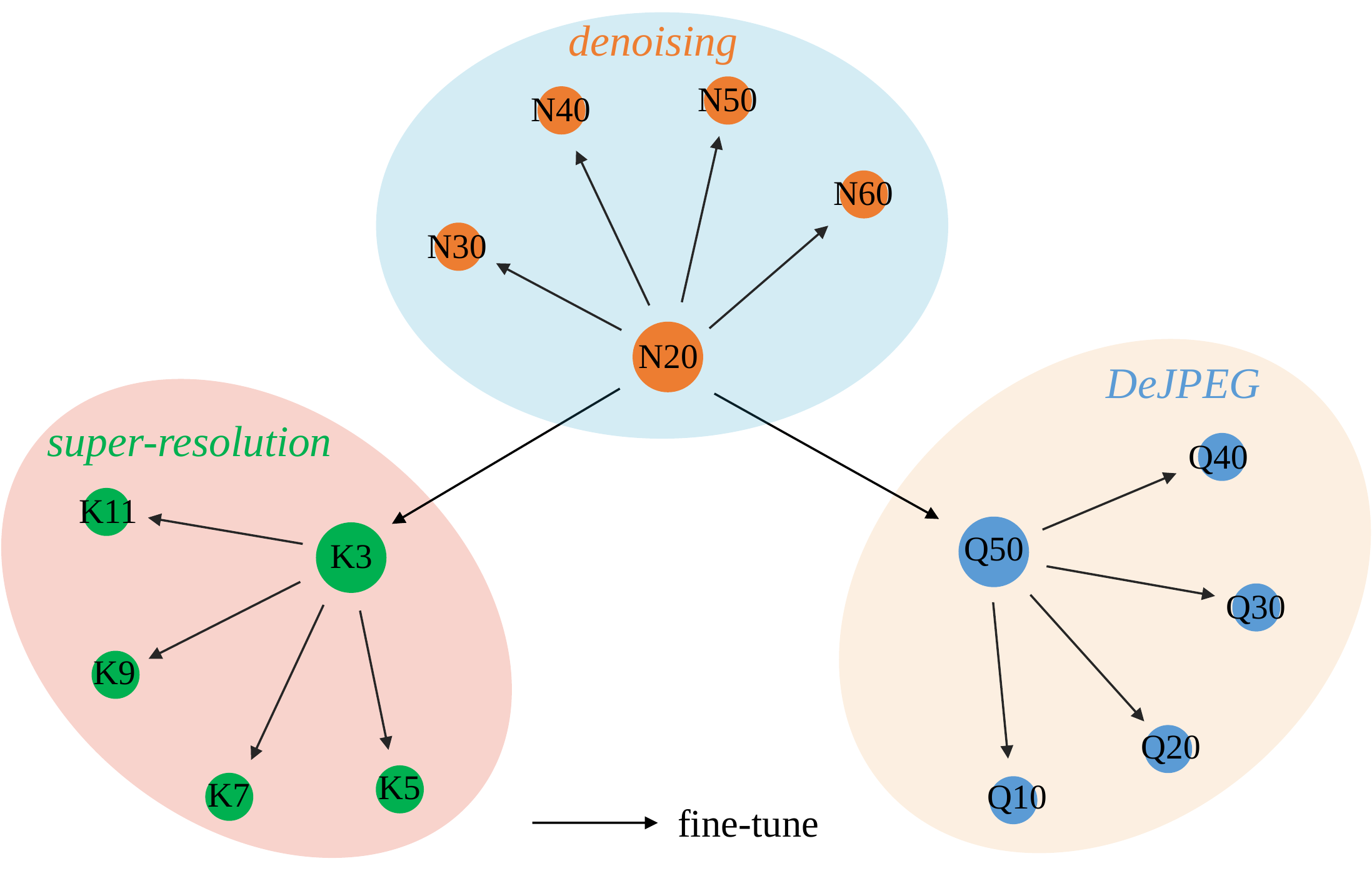}
		\caption{Each dot represents a model.  Lines with arrows denote fine-tuning. For instance, the 
		super-resolution (K3) model is fine-tuned from the denoising (N20) model. 
		}
		\label{fig:model_finetune}
	\end{center}
	\vspace{-0.7cm}
\end{figure}

We obtain all the models from one pre-trained model by fine-tuning. The fine-tuning maps are depicted in 
Fig.~\ref{fig:model_finetune}. Each dot in the figure represents a model. Lines with arrows denote fine-tuning. For 
instance, the super-resolution (K3) model is fine-tuned from the denoising (N20) model. 

\begin{figure*}[th]
	\begin{center}
		\includegraphics[width=\linewidth]{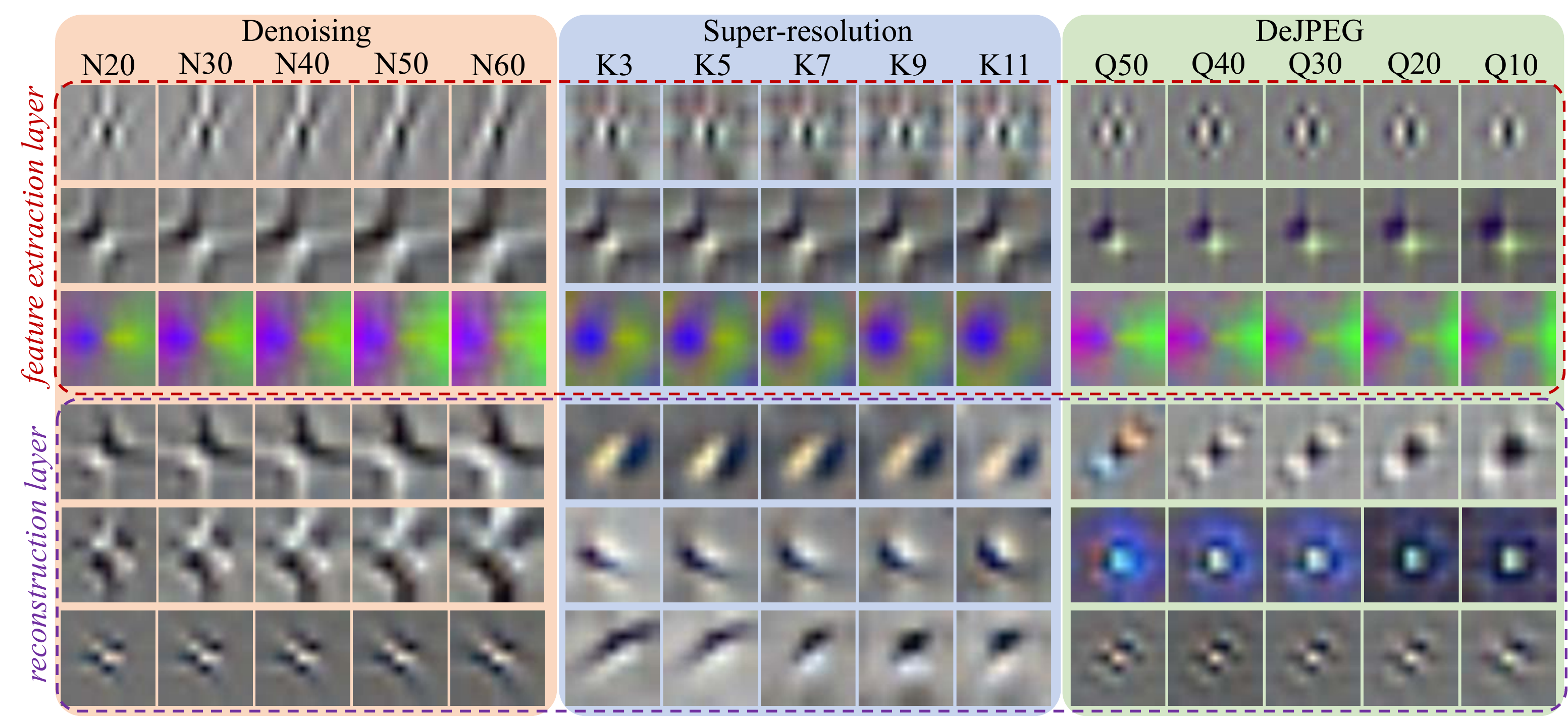}
		\caption{Filter visualization examples of the \textit{feature extraction} and \textit{reconstruction} layers 
			for denoising, super-resolution and DeJPEG tasks. 1) The learned filters for different 
			distortion levels under the same degradation exhibit continuous changes. 2) Filters 
			for different image distortions (\eg, denoising and super-resolution) capture the special 
			characteristics of their own tasks, representing different filter patterns, especially in the 
			\textit{reconstruction} layer.
		}
		\label{fig:filter_denoise_sr_dejpeg}
	\end{center}
	\vspace{-0.3cm}
\end{figure*}

\begin{figure*}[th]
	\begin{center}
		\includegraphics[width=\linewidth]{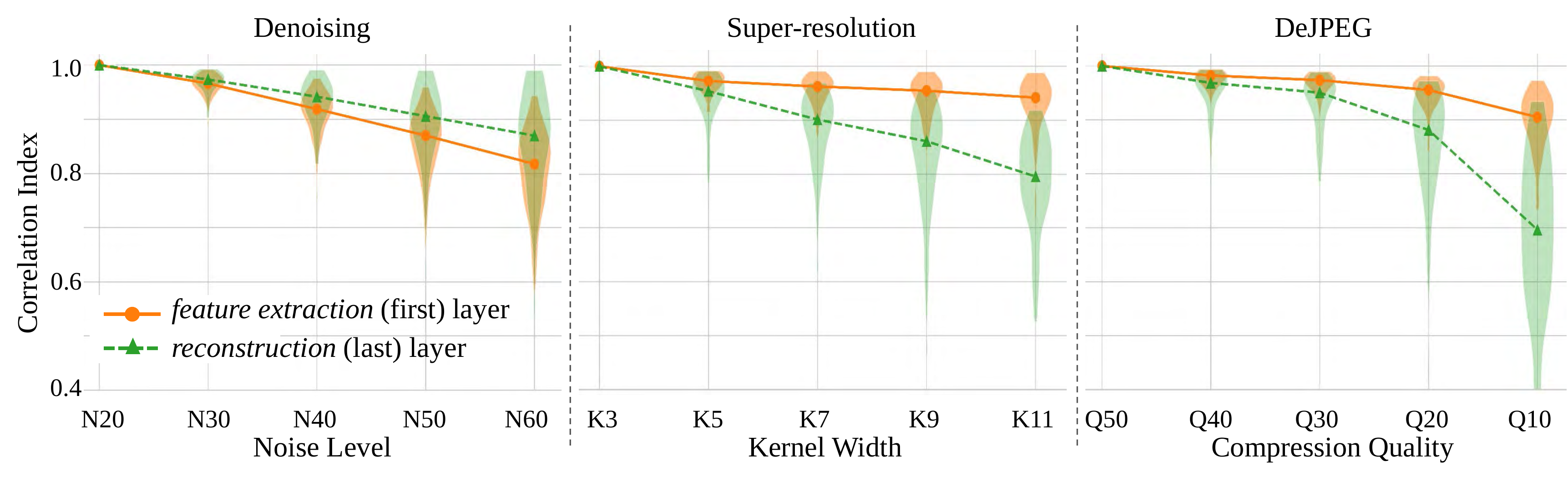}
		\caption{Filter correlation index curves of the \textit{feature extraction} and \textit{reconstruction} layers 
			for denoising, super-resolution and DeJPEG tasks. The curves present the median of correlation 
			indexes and the correlation distributions are also plotted. These curves show that the learned filters for 
			different distortion levels under the same distortion exhibit continuous changes. 
		}
		\label{fig:filter_cor}
	\end{center}
	\vspace{-0.3cm}
\end{figure*}
 
Fig.~\ref{fig:filter_denoise_sr_dejpeg} visualizes several filter examples of the feature extraction and 
reconstruction layers for denoising, super-resolution and DeJPEG tasks.
We can found that: 1) Under the constraint of fine-tuning, the learned 
filters for related tasks exhibit continuous changes. This phenomenon is observed in all the tasks, including 
denoising, super-resolution and DeJPEG.
2) Except the similarity, we further see that filters for different image 
distortions (\eg, denoising and 
super-resolution) capture the special characteristics of their own tasks, representing different filter patterns, 
especially in the reconstruction layer.
Thus, if two tasks are far way from each other, the weak correlation of filters 
could result in unsatisfying, even meaningless interpolated results.
The definition of task distances and the application boundaries of DNI are still open questions. Our analyses focus on 
related tasks of different degradations under the same distortion. It is noteworthy that DNI is capable of dealing with 
lot of tasks for continuous imagery effect transition and its broad applications indicate that related tasks with close 
``distances'' are common and practical in real-world scenarios.

\begin{figure*}[th]
	\centering
	\vspace*{2mm}\begin{subfigure}[b]{0.95\linewidth}
		\includegraphics[width=\linewidth]{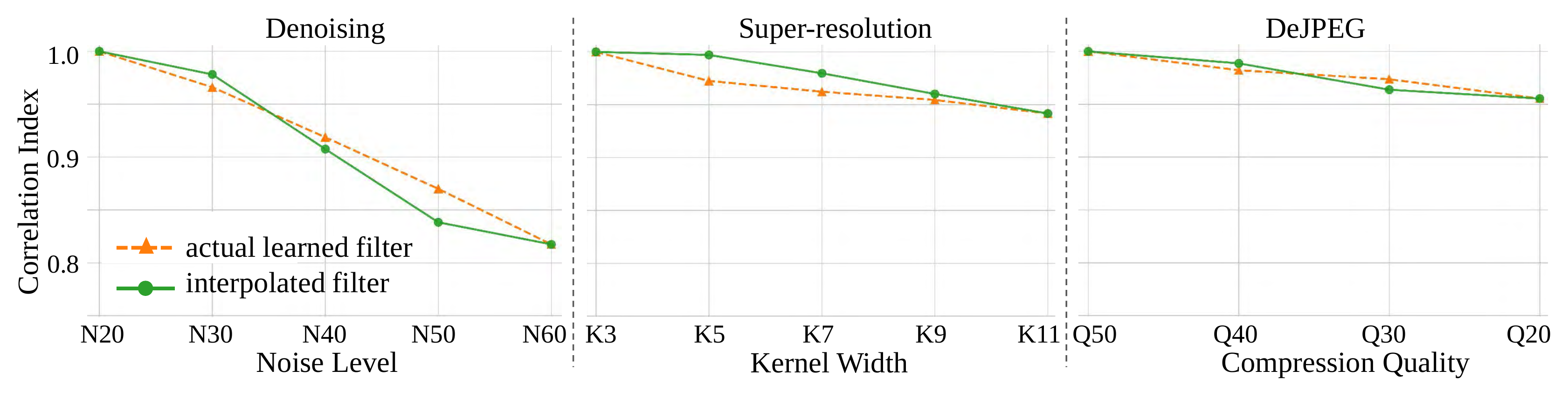}
		\vspace*{-7mm}\caption{The \textit{feature extraction} layer (first layer).}
	\end{subfigure}
	\\
	\vspace*{2mm}\begin{subfigure}[b]{0.95\linewidth}
		\includegraphics[width=\linewidth]{figs/interp_filter_cor1.pdf}
		\vspace*{-7mm}\caption{The \textit{reconstruction} layer (last layer).}
	\end{subfigure}
	\vspace*{-2mm}
	\caption{The correlation curves for actual learned filters and interpolated filters are very 
		close for both the \textit{feature extraction} and \textit{reconstruction} layers, indicating that the 
		interpolated filters could fit learned filters well.
		The similar observations could be found on denoising, super-resolution and DeJPEG tasks.
	}
	\label{fig:interp_filter_cor}
	\vspace*{0mm}
\end{figure*}

We also calculate the filter correlation indexes and plot their correlation distribution. The curves of the 
\textit{feature extraction} and \textit{reconstruction} layers for denoising, super-resolution and DeJPEG tasks are 
shown in Fig.~\ref{fig:filter_cor}. We again observe continuous changes of learned filters for different distortion 
levels under the same distortion.

We then show that the interpolated filters fit the learned filters well for different layers in the networks, and 
also for different tasks including denoising, super-resolution and DeJPEG.
We perform linear interpolation between the filters from the two ends of degradation levels (\eg, the N20 and N60
models for denoising; the K3 and K11 models for super-resolution).
With optimal interpolation coefficients $\alpha$, the interpolated filters could visually fit those learned filters for 
all the three 
tasks, as shown in Fig.~\ref{fig:interp_filter_cor}. The observations could 
be held for both the feature extraction and reconstruction layers. 

The optimal $\alpha$ is obtained through the final performance of the interpolated 
network. Specifically, we perform DNI with $\alpha$ from 0 to 1 with an interval of 0.05. The best $\alpha$ for
each degradation level is selected based on the highest PSNR on the
test dataset.

As the network goes deeper, the non-linearity increases and the network behaviors become more complicated. We provide a 
preliminary analysis of deeper denoising network in the main paper, where the observations are consistent with our 
conclusion.
For other tasks, such as image style transfer or image translation, the investigation from the filter perspective 
becomes more sophisticated, since these tasks cannot be defined continuously like the continuous degradation levels
in image restoration. 
The exploration of inherent filter correlations and the in-depth reason why DNI works are worth investigated in the
future work.

\begin{figure}[h]
	\begin{center}
		\includegraphics[width=\linewidth]{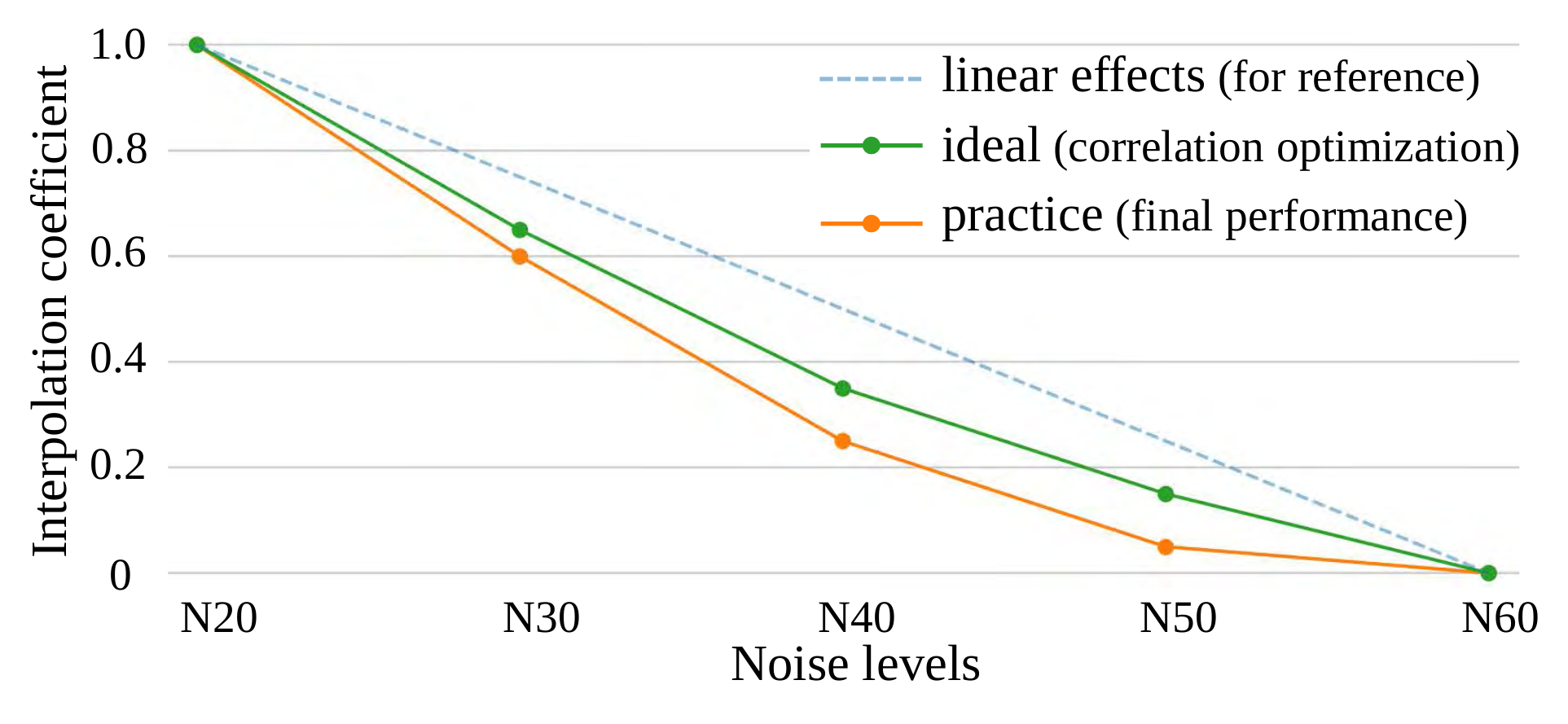}
		\caption{To achieve ``linear'' transition effects, the coefficients for each level are non-linear both from 
			ideal analyses (optimizing the filter correlation) and practical performance (obtaining from the final 
			PSNR.) 
		}
		\label{fig:interp_nonlinear}
	\end{center}
\end{figure}

\vspace{3mm}
\noindent\textbf{The non-linearity between interpolation coefficients and effects.}
In our analyses, in order to realize ``linear effects'' of outputs, (\ie obtaining denoising models dealing with 
N30, N40 and N50), the different interpolation coefficients do not present a linear relationship. The practical 
optimal $\alpha$, obtained through the final performance of the interpolated 
network, for each noise level is shown in Fig.~\ref{fig:interp_nonlinear} (orange curve).
One reason is that the effects of filter interpolation and the output effects are inherently not linear.
We can also obtain the optimal $\alpha$ by optimizing the filter correlation index with models trained for each 
noise levels.
A similar non-linear trend as the practical curve could be observed (Fig.~\ref{fig:interp_nonlinear}, green curve).
We suspect that the extra gap between the green and orange curves may be from the non-linearity in networks.

The non-linear sampling of $\alpha$ for ``linear'' transition effects could be observed in many DNI applications. 
However, it does not 
influence its extensive applications. A solution is to control the sampling density of $\alpha$, simply resulting in 
``linear'' transition effects. 

\section{Implementation Details}\label{sec:implementation_details}
In this section, we provide the implementation details of DNI and the fine-tuning strategy for several applications in the main 
paper.

\noindent\textbf{Balance MSE and GAN effects in super-resolution.}
The MATLAB bicubic kernel with a scaling factor of 4 is adopted as the down-sampling kernel.
We first train a super-resolution model with MSE loss~\cite{timofte2017ntire}, which tends to produce over-smooth 
images.
We then fine-tune it with GAN loss and perceptual loss~\cite{ledig2017photo}, obtaining results with vivid details yet 
accompanied with unpleasant artifacts.
DNI is applied in these two models. 
We use dense blocks~\cite{huang2016densely} as the network architecture.

\noindent\textbf{Image-to-image Translation.}
In image-to-image translation, we use the popular CycleGAN~\cite{zhu2017unpaired} to explore the broad applications of 
DNI\footnote{We use the official released 
codes: \url{https://github.com/junyanz/pytorch-CycleGAN-and-pix2pix}.}.
In the original CycleGAN, there are two networks $G_A$ and $G_B$ to learn a mapping and its inverse mapping, 
respectively. For 
instance, in order to translate landscape photos to paintings, $G_A$ learns the mapping from paintings 
to photos while $G_B$ learns an inverse mapping from photos to painting.

We use slightly different settings for different applications. For \textit{mixing various painting styles}, we first 
train a 
CycleGAN model for translating photos to paintings with Van Gogh's style, \ie, $G_A^{Van Gogh}$ is used for turning Van 
Gogh paintings into photos, and $G_B^{Van Gogh}$ for translating photos into paintings with Van Gogh's style. 
We then fine-tune these two networks $G_A^{Van Gogh}$ and $G_B^{Van Gogh}$ (together with the discriminators) to models 
with another painter style, such as Monet, and obtain $G_A^{Monet}$ and $G_B^{Monet}$. 

During inference, our aim is to translate a landscape photo to paintings with various styles, even the mixtures of 
several famous painters. Thus, we only keep the $G_B^{Van Gogh}$ and $G_B^{Monet}$ and perform DNI on these 
two networks.

For \textit{day-to-night application}, we first train a CycleGAN model as usual, \ie, $G_A$ is used for translating a 
day photo to a night one while $G_B$ translates the night photo to the day one. Note that the CycleGAN model is only 
able to translate between two states and cannot produce a series of images with smooth transitions from day to night.
We then fine-tune the whole pre-trained model with identity mapping. Specifically, we remove the GAN loss, adopt a 
$10\times$ identity loss and keep the cyclic loss in the CycleGAN framework. Thus, the fine-tuned network 
$G_A^{identity}$ 
always outputs identical results, \ie, it receives a day photo and outputs the same day photo.

DNI is then applied in $G_A$ and $G_A^{identity}$. With an appropriate coefficient, the interpolated network is able to 
produce images with arbitrary effects between day and night. The same operations are also employed in the 
\textit{deep-to-shallow depth of filed} application.

\noindent\textbf{Style Transfer.}
We use the PyTorch example codes for style 
transfer\footnote{\url{https://github.com/pytorch/examples/tree/master/fast_neural_style}.}.
We first train a model for one certain style and then fine-tune it for another style. DNI is performed on these two 
models. For the stroke factor, we fine-tune the pre-trained model for a style image with different size.
In order to balance the content and style, we fine-tune 
the pre-trained model with smaller style loss, resulting in almost identity mapping.

We note that DNI is generic for several controllable factors in style transfer, such as styles, strokes, and the balance of 
content and style. The proposed DNI could be also applied to more advanced models~\cite{dumoulin2016learned,huang2017arbitrary}
that address multiple or arbitrary style transfer, resulting in more diverse outputs.

\section{More Applications and Analyses}\label{sec:more_applications_and_analyses}

\subsection{Extend Restoration Models to Unseen Distortion Levels}

\begin{figure*}[th]
	\begin{center}
		\includegraphics[width=\linewidth]{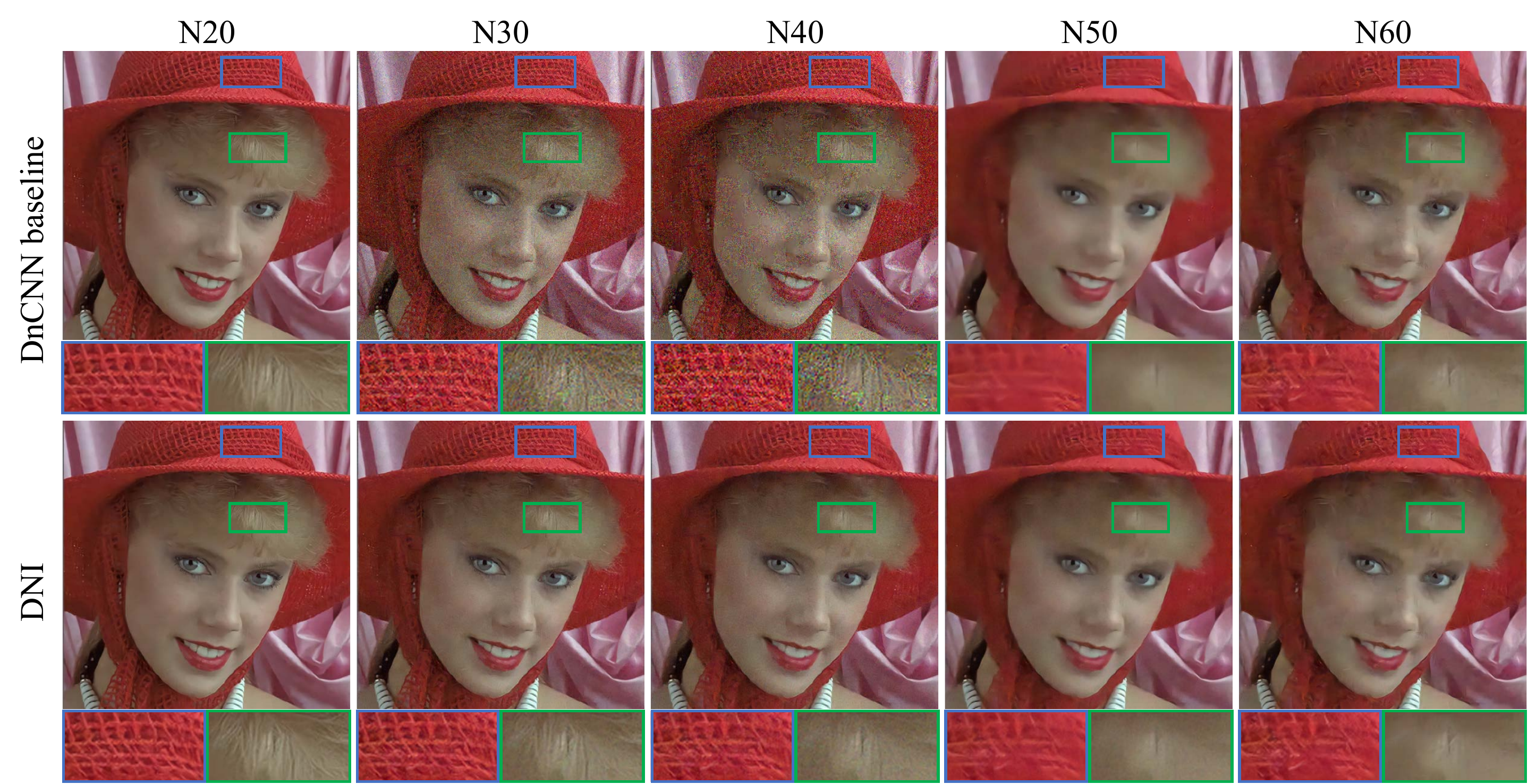}
		\caption{Extending denoising models to unseen noise levels. The baseline model trained with N20 and N60 data is 
			incapable of removing unseen noise (\eg, N30 and N40), while our method  could deal with these 
			unseen scenes. 
			(\textbf{Zoom in for best view})}
		\label{fig:denoise_interp}
	\end{center}
	\vspace*{-5mm}
\end{figure*}

In Sec.~\ref{sec:filter_analyses}, we reveal the inherent correlation of learned filters for a series of related 
tasks. Here, we take advantage of this observation to extend restoration models to deal with unseen distortion 
levels. We take the denoising task for example and it could also be applied to other restoration tasks, \eg, 
super-resolution with different down-sampling kernels. 

We employ an extreme case, where only the data with noise level N20 and N60 are available during training 
and we expect the trained models are able to handle arbitrary noise levels in the middle during testing. In particular, 
we test unseen N30, N40, and N50 noise levels.
The upper bound for each level is the model trained with its corresponding data. The baseline is the model trained with 
both the N20 and N60 data.
We adopt DnCNN~\cite{zhang2017beyond} as the denoising model\footnote{We use the official released 
codes:\url{https://github.com/cszn/DnCNN} (PyTorch version).} and use two variants, with and without BN. 

We first train a model using N20 data and then fine-tune it with N60 data. DNI is then performed with interpolation 
coefficients $\alpha \in [0, 1]$ with an interval of 0.1. 
The best result for each noise level is selected among these interpolated models (the chosen $\alpha$ for each level 
is shown in Tab.~\ref{tb:DnCNN_denoise}). 
We note that the selection is simple with a few trials due to the smooth transition. For automatic denoising, we could 
further train a shallow network to regress a proper $\alpha$ according to the noisy input.

We evaluate DNI using the LIVE1~\cite{sheikh2005live} dataset. Quantitative results with the PSNR metric
and qualitative results are shown in Tab.~\ref{tb:DnCNN_denoise} and Fig.~\ref{fig:denoise_interp} respectively.
The baseline model is incapable of removing unseen noise (\eg, N30 and N40 in Fig.~\ref{fig:denoise_interp}), leading 
to drastic drops in performance. However, our method could deal with those unseen scenes, even approaching 
to the upper bound. (Note that the upper bound observes the corresponding data.)

Though our DNI with BN can outperform its corresponding baseline, there is still a little drop compared with that 
without BN. The BN effects for DNI are still an open question and a better interpolation scheme needs to be explored 
for normalization layers. 
We also note that this application demonstrates that it is worth exploiting the underlying relations of learned 
filters to further extend the ability and practicality of existing models.

\begin{figure*}[th]
	\begin{center}
		\includegraphics[width=\linewidth]{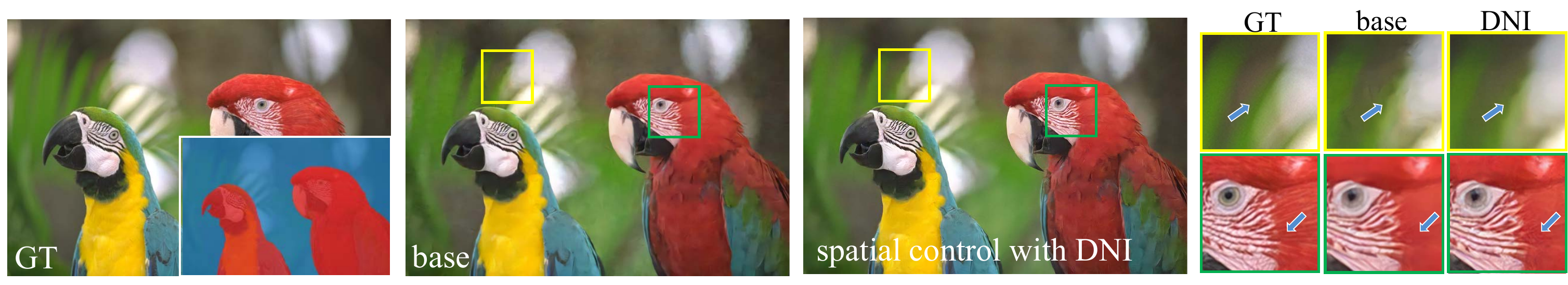}
		\caption{Spatial control for adjustable denoising.
			With a mask, different denoising strengths are applied separately for the foreground and the background. 
			(\textbf{Zoom in for best view})}
		\label{fig:denoise_spatial_control}
	\end{center}
	\vspace*{-5mm}
\end{figure*}

\begin{table}[htbp]
	\centering
	\caption{The average denoising results of PSNR (dB) on the LIVE1 test dataset. Unseen noise levels are denoted with 
		*. Note that the upper bound have seen the corresponding data.}
	\label{tb:DnCNN_denoise}
	\tabcolsep=0.16cm
	\begin{tabular}{cc|ccccc}
		\hline
		& Noise level  & N20  & N30*  & N40*  & N50*  & N60  \\ \hline \hline
		
		\multicolumn{1}{c|}{\multirow{4}{*}{\rotatebox{90}{w/o BN}}}
		& Upper bound  & 32.38  & 30.39  & 29.01  & 27.98  & 27.18  \\
		\multicolumn{1}{c|}{}
		& Baseline  & 32.36  & 23.86  & 21.90  & 27.34  & 27.14  \\
		\multicolumn{1}{c|}{}
		& \textbf{DNI (ours)} & \textbf{32.38} & \textbf{29.84} & \textbf{28.28} & \textbf{27.67} & \textbf{27.18} \\
		\multicolumn{1}{c|}{}
		& $\alpha$  & 1  & 0.7  &  0.4  & 0.1  & 0  \\ \hline \hline
		
		\multicolumn{1}{c|}{\multirow{4}{*}{\rotatebox{90}{w/ BN}}}
		& Upper bound  & 32.49  & 30.48  & 29.09  & 27.96  & 27.25  \\
		\multicolumn{1}{c|}{}
		& Baseline  & 32.42  & 24.42  & 26.58  & 27.44  & 27.21  \\
		\multicolumn{1}{c|}{}
		& \textbf{DNI (ours)} & \textbf{32.49} & \textbf{29.46} & \textbf{28.08} & \textbf{27.66} & \textbf{27.25} \\
		\multicolumn{1}{c|}{}
		& $\alpha$  & 1  & 0.6  & 0.3  & 0.1  & 0  \\ \hline
	\end{tabular}
	\vspace*{-5mm}
\end{table}

\subsection{Spatial Control for Adjustable Denoising}
In the main paper, we emphasize the importance of adjustable denoising strength and show the ability of DNI to satisfy 
the needs.
Here, we further present an application of spatial control for adjustable denoising. For the DSLR photos with shallow 
depth-of-filed, the background is usually blurred while the foreground contains rich details. We can easily separate 
them with a mask and adopt different denoising strengths respectively, obtaining better visual quality.
From Fig.~\ref{fig:denoise_spatial_control}, we can see that with adjustable denoising realized by DNI, the blurry area 
is more smooth without artifacts, while there are rich details in texture regions.

Apart from the denoising task, the adjustments with DNI can also be applied to other image restoration tasks, \eg, 
super-resolution with different down-sampling kernels and JPEG artifacts removal with different compression qualities.

\subsection{Multi-ends DNI}
\begin{figure*}[th]
	\centering
	\vspace*{2mm}\begin{subfigure}[b]{\linewidth}
		\includegraphics[width=\linewidth]{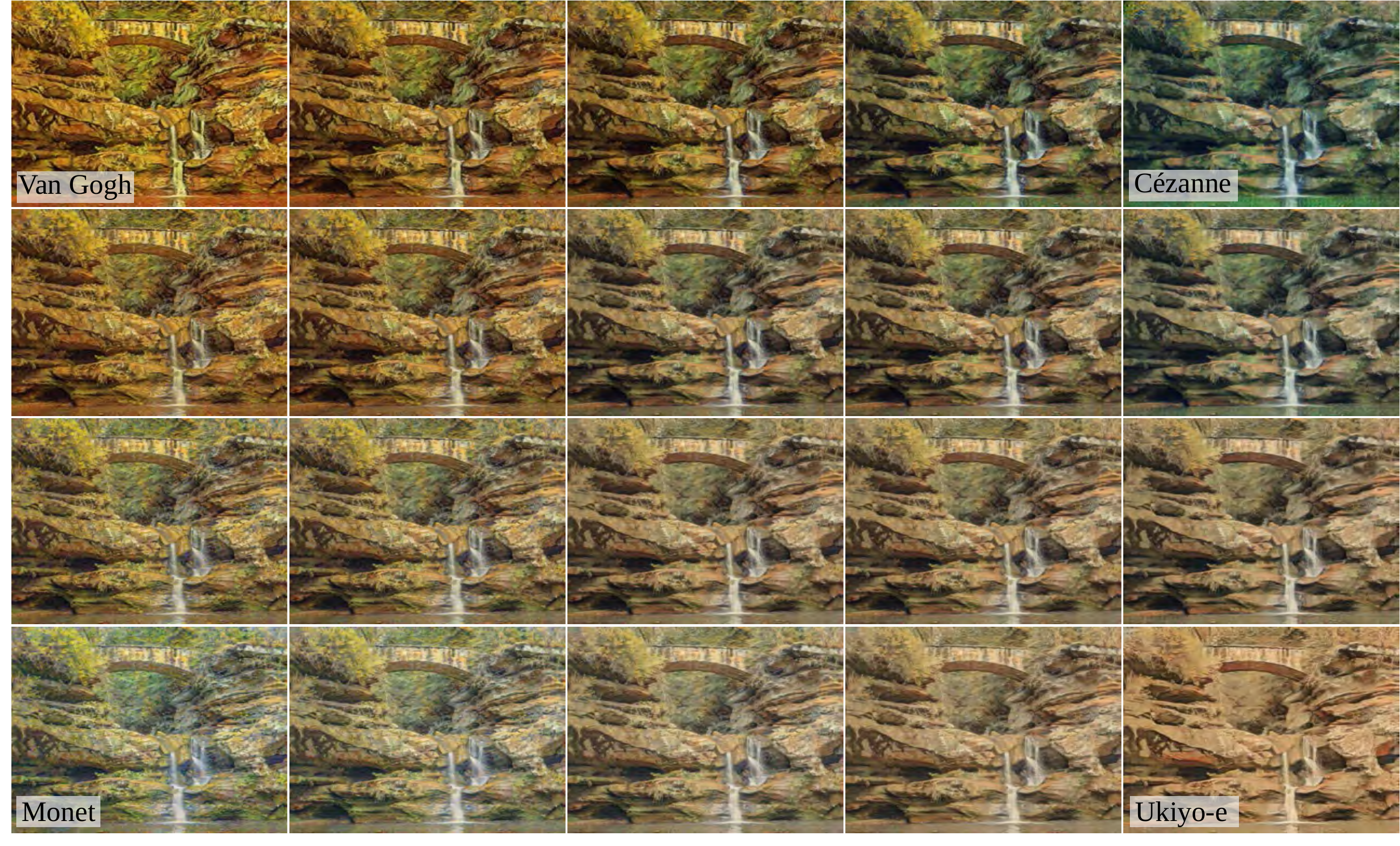}
	\end{subfigure}
	\\
	\vspace*{2mm}\begin{subfigure}[b]{\linewidth}
		\includegraphics[width=\linewidth]{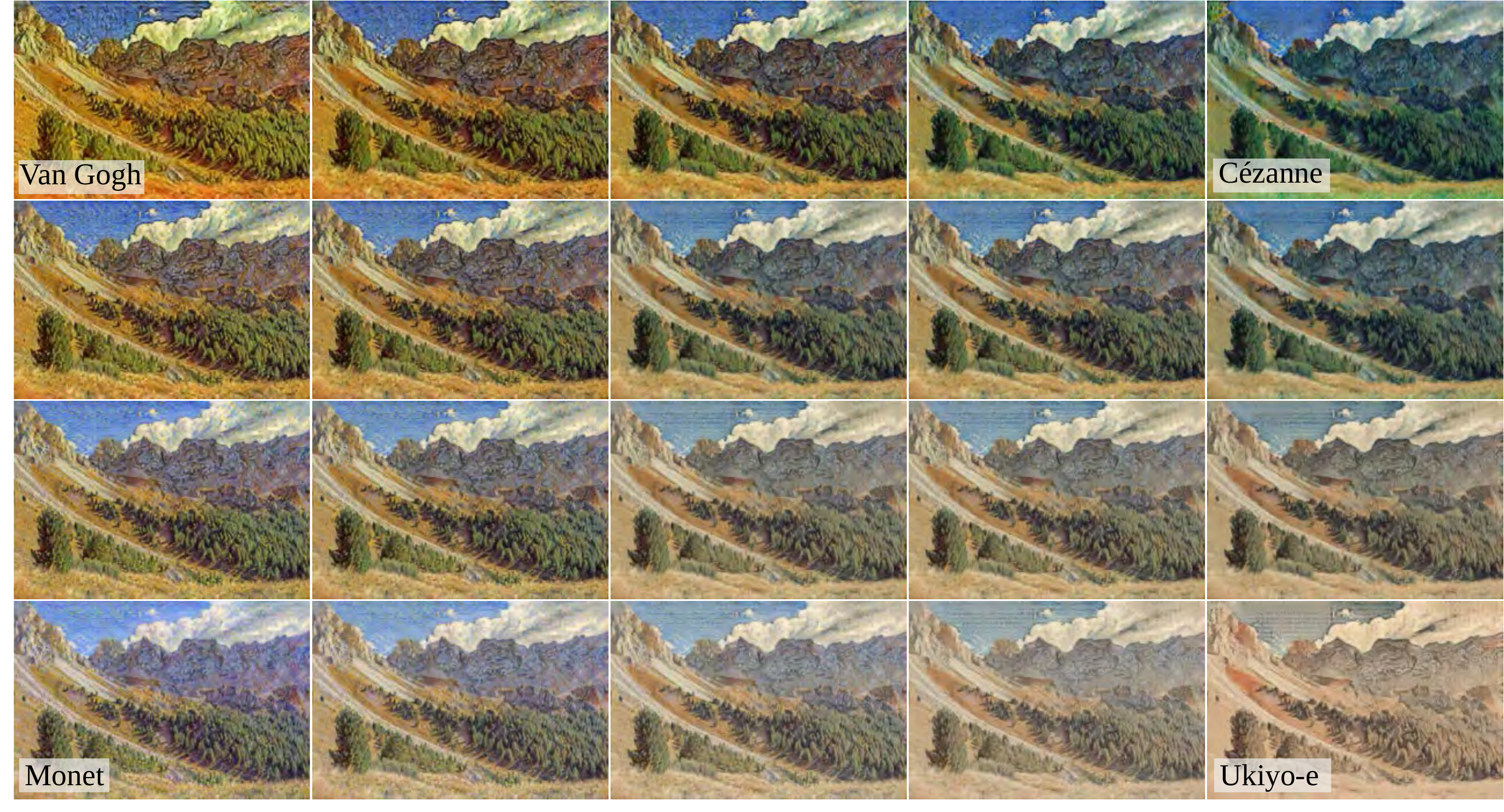}
	\end{subfigure}
	\vspace*{-5mm}
	\caption{(Two examples) Translating landscape photos to paintings with various 
		styles -- Van Gogh, C\'ezanne, Monet and Ukiyo-e. By adjusting the interpolated coefficients, richer and more
		diverse effects with continuous transitions could be realized.
	}
	\label{fig:painting_multi_ends}
	\vspace*{0mm}
\end{figure*}

A general form of DNI is also capable of interpolating more than two networks. 
Fig.~\ref{fig:painting_multi_ends} shows two examples of translating landscape photos to paintings with various 
styles -- Van Gogh, C\'ezanne, Monet and Ukiyo-e. By adjusting the interpolated coefficients, richer and more diverse 
effects with continuous transitions could be realized.

\begin{figure*}[h]
	\begin{center}
		\includegraphics[width=0.98\linewidth]{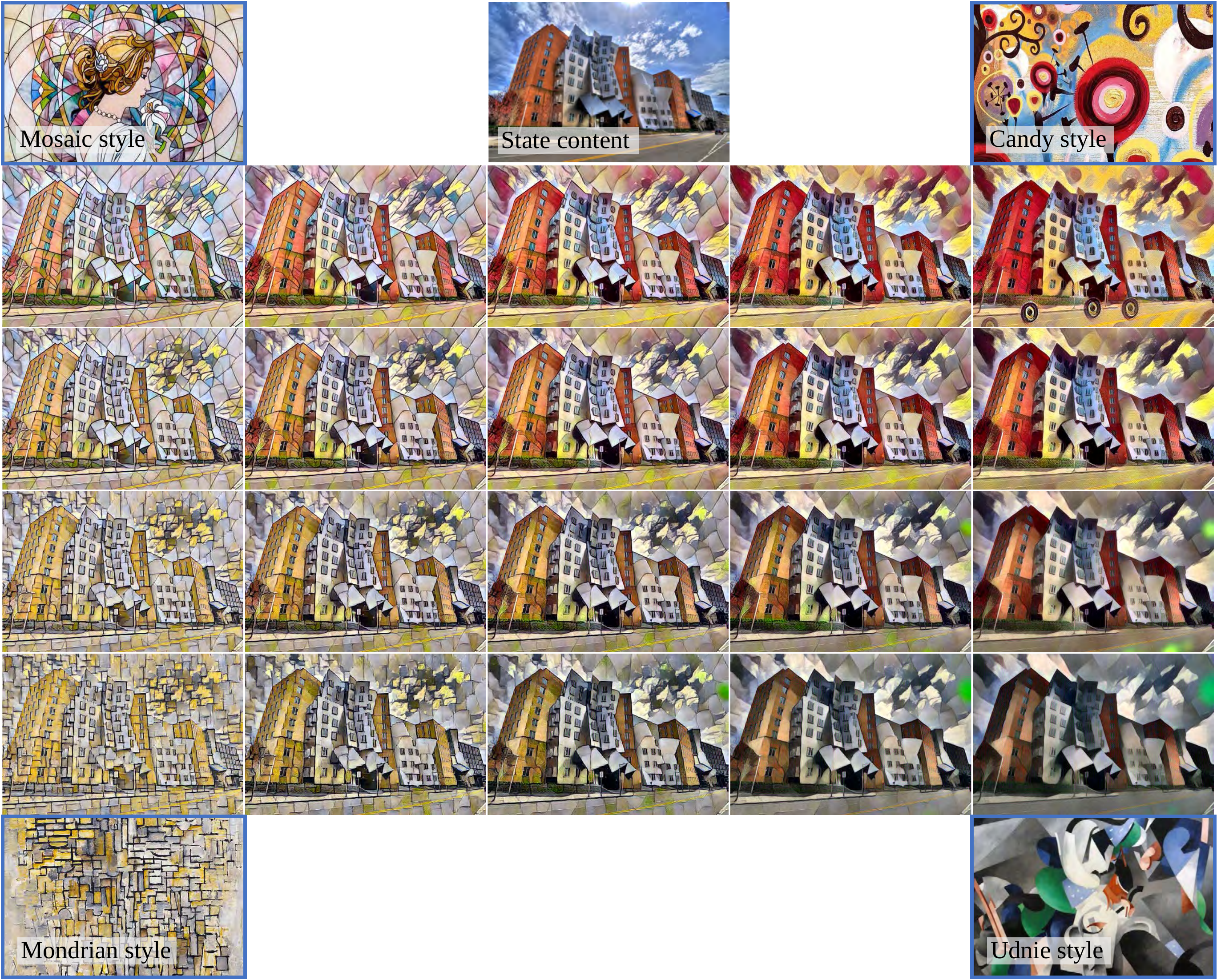}
		\caption{Image style transfer among different styles -- Mosaic style, Candy style, Mondrian style and Udnie 
		style. It generates diverse and new styles, meeting users' various aesthetic flavors. 
		}
		\label{fig:styles}
	\end{center}
	\vspace*{-4mm}
\end{figure*}

Another example of image style transfer in Fig.~\ref{fig:styles} presents the ability of DNI to transfer among 
different styles -- Mosaic style, Candy style, Mondrian style and Udnie style. It generates diverse and new styles, 
meeting users' various aesthetic flavors. 

